\pdfoutput=1

\documentclass[11pt]{article}

\usepackage[final]{acl}

\usepackage{times}
\usepackage{latexsym}
\usepackage{booktabs}
\usepackage{hyperref} 
\usepackage{tcolorbox}
\usepackage{subcaption}
\usepackage{array}
\usepackage{xcolor}
\usepackage{soul}
\usepackage{multicol,multirow}
\usepackage{todonotes}
\usepackage{xspace}
\definecolor{mygold}{HTML}{eeba0a}
\definecolor{mygrey}{HTML}{bac8ca}
\definecolor{mylightgreen}{HTML}{D1F694}
\definecolor{myblue}{HTML}{30C0F0}
\definecolor{myorange}{HTML}{F5AF22}
\definecolor{myred}{HTML}{FF576A}
\definecolor{darkred}{HTML}{91270F}
\definecolor{darkgreen}{HTML}{5E893E}
\definecolor{lightblue}{HTML}{b6ecfc}
\sethlcolor{mylightgreen}
\usepackage{makecell}
\usepackage{longtable}
\usepackage{amssymb}
\usepackage{pifont}
\newcommand{\xmark}{\textcolor{darkred}{\ding{55}}}%
\newcommand{\cmark}{\textcolor{darkgreen}{\ding{51}}}

\definecolor{lightpink}{rgb}{1.0, 0.8, 0.9}

\newcommand{\digitbf}[1]{%
  \colorbox{lightblue}{\textbf{#1}}%
}
\newcommand{\digitorange}[1]{%
  \colorbox{myorange}{\textbf{#1}}%
}

\usepackage[T1]{fontenc}

\usepackage[utf8]{inputenc}

\usepackage{microtype}

\usepackage{inconsolata}

\usepackage{graphicx}

\title{Is this chart lying to me? \\ Automating the detection of misleading visualizations}

\author{
Jonathan Tonglet\thanks{These authors contributed equally to this work.}$^{1,2,3}$, Jan Zimny\footnotemark[1]$^{1,2}$, Tinne Tuytelaars$^{2}$, Iryna Gurevych$^{1}$
\\
        \textsuperscript{1} Ubiquitous Knowledge Processing Lab (UKP Lab), Department of Computer Science, \\ TU Darmstadt and National Research Center for Applied Cybersecurity ATHENE\\ 
\textsuperscript{2} Department of Electrical Engineering, KU Leuven\\
\textsuperscript{3} Department of Computer Science, KU Leuven\\
\href{https://www.ukp.tu-darmstadt.de}{www.ukp.tu-darmstadt.de}
}

\begin{document}
\maketitle
\begin{abstract}
Misleading visualizations are a potent driver of misinformation on social media and the web. By violating chart design principles, they distort data and lead readers to draw inaccurate conclusions.  Prior work has shown that both humans and multimodal large language models (MLLMs) are frequently deceived by such visualizations. Automatically detecting misleading visualizations and identifying the specific design rules they violate could help protect readers and reduce the spread of misinformation. However, the training and evaluation of AI models has been limited by the absence of large, diverse, and openly available datasets.
In this work, we introduce Misviz, a benchmark of 2,604 real-world visualizations annotated with 12 types of misleaders. To support model training, we also create Misviz-synth, a synthetic dataset of 57,665 visualizations generated using Matplotlib and based on real-world data tables. We perform a comprehensive evaluation on both datasets using state-of-the-art MLLMs, rule-based systems, and image-axis classifiers. Our results reveal that the task remains highly challenging. We release Misviz, Misviz-synth, and the accompanying code.\footnote{\href{https://github.com/UKPLab/acl2026-misviz}{github.com/UKPLab/acl2026-misviz}}

\end{abstract}

\section{Introduction}

Misleading visualizations are charts that distort the underlying data, typically by violating design principles, leading readers to draw inaccurate conclusions \citep{tufte1983visual,10.1145/2702123.2702608,10.1145/3380851.3416762,10.1145/3313831.3376420,lo2022misinformed,10.1145/3544548.3580910,10670488}.  While many arise from unintentional design errors, misleading visualizations are also deliberately crafted by malicious actors to spread disinformation and manipulate public understanding, especially during crises such as the COVID-19 pandemic, where misleading charts circulated widely on social media \citep{2017-blackhatvis,10.1145/3544548.3580910,10.1111/bjso.12787}. Prior work has shown that both humans \citep{6876023,10.1145/2702123.2702608,10.1145/3233756.3233961,YANG2021298,10.1145/3544548.3581406,rho2023various} and MLLMs \citep{10670574,chen2025unmaskingdeceptivevisualsbenchmarking,pandey2025benchmarkingvisuallanguagemodels,tonglet2025protectingmultimodallargelanguage} are easily deceived by such visualizations in question-answering tasks. 

\begin{figure*}[!ht]
    \centering
    \includegraphics[width=\linewidth]{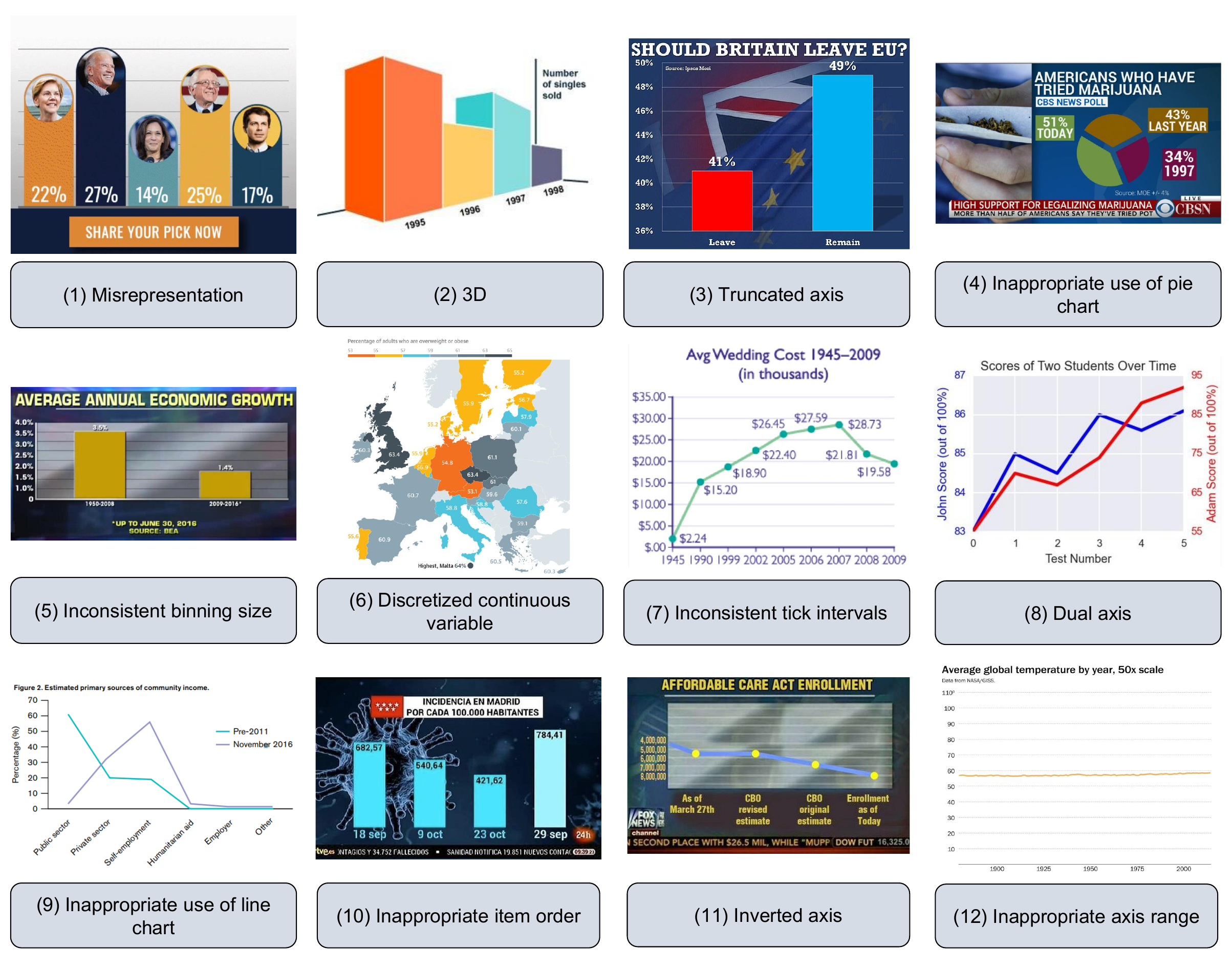}
    \caption{Examples of the 12 types of misleaders included in Misviz. Appendix \ref{sec:solution} explains how these visualizations misrepresent their underlying data table.}
    \label{fig:misleaders}
\end{figure*}

The deceptive features in these charts, or \textit{misleaders} \citep{10.1145/3544548.3580910}, often reside in subtle details easily missed by readers, such as axis tick intervals. Furthermore, misleaders are highly diverse: the latest taxonomies identify over 70 distinct types spanning a wide range of chart types, including bar charts, pie charts, and choropleth maps \citep{lo2022misinformed,10670488}. In some cases, multiple misleaders affect the same visualization \citep{lo2022misinformed}. Figure~\ref{fig:misleaders} shows 12 real-world examples of misleading visualizations. 

Automatically classifying whether a visualization is misleading and identifying which misleaders affect it, if any, can enable timely warnings to chart designers and readers and help prevent the spread of misinformation. This task is framed as a multi-label classification problem. While early work relied on rule-based systems called linters \citep{10.1111/cgf.13975,10.1145/3491102.3502138}, recent studies have explored the use of MLLMs \citep{10679256,10771149}. However, these approaches were evaluated on distinct datasets, which are either small or closed-access, limiting comparability and hindering progress.

In this work, we introduce Misviz, a large, diverse, and open benchmark comprising 2,604 real-world visualizations spanning 12 types of misleaders. It reflects scenarios in which detection models could flag visualizations published on the web. In Misviz, 70\% of the visualizations contain up to three misleaders, while the remaining 30\% are non-misleading. To support model training, we also release Misviz-synth, a synthetic dataset generated with Matplotlib \citep{Hunter:2007} using real-world data tables. Misviz-synth reflects scenarios in which detection models assist chart designers in identifying misleaders unintentionally introduced into their charts. The dataset includes not only the visualizations but also their underlying data tables, Python code snippets, and axis metadata, enabling the training of chart de-rendering models.

We conduct extensive experiments with three approaches: (a) state-of-the-art MLLMs, (b) a new rule-based linter that inspects axis metadata for design rule violations, and (c) new classifiers that take the visualization alone or in combination with the axis metadata as input. For (b) and (c), we fine-tune DePlot \citep{liu-etal-2023-deplot} to extract axis metadata as an intermediate step. Our experiments address the following research questions (RQs). \textbf{RQ1}: Which type of model performs best on real-world or synthetic instances? \textbf{RQ2}: Can detection models trained on synthetic instances generalize to real-world cases? \textbf{RQ3}: Can axis extraction models trained on synthetic instances generalize to real-world cases? 

Our results show that MLLMs perform best on real-world visualizations, while linters and image-axis classifiers outperform them on synthetic ones,
benefiting from the availability of training data for both axis extraction and misleader detection. While
the fine-tuned DePlot can extract axes from Misviz-synth, it does not generalize well to Misviz, reducing the performance of the linter and classifier.

In summary, our contributions are as follows:
(1) We introduce Misviz and Misviz-synth, the first large-scale open datasets for misleading visualization detection. (2) We propose a new linter and a new classification method that combines image and extracted axis metadata as input. (3) We conduct a comprehensive evaluation and error analysis, highlighting the strengths and weaknesses of each method and identifying directions for future work.

\section{Related work}

\begin{table*}
  
  \centering
  \resizebox{\textwidth}{!}{ %
  \begin{tabular}{lcccccccc}
    \hline       
    Dataset     & Instances & Misleader types & Chart types & \% non-misleading &     Open access & Real-world & Multi-label & Axes, table, code \\
    \hline
    MISCHA-QA \citep{mischaqa} & 8,201 & 4 & 3 & 49  &  \cmark & \xmark  & \xmark  & \xmark \\
    DCDM \citep{10.1007/978-3-031-98414-3_7} &  24,480 & 5 & 3 & 51 &  \cmark & \xmark  & \xmark  & \xmark  \\
    \citet{10771149} - \textit{design misleaders} & 1,460 & 7 & > 5 & 50 & \xmark & \cmark &  \xmark  & \xmark\\
    \citet{10679256}  & 150  &  21  & > 5 & 16 &   \cmark   &  \cmark  & \xmark  & \xmark\\
    Misvisfix \citep{das2025misvisfixinteractivedashboarddetecting} & 450  &  74  & > 5 & 20 &   \cmark   &  \cmark  & \cmark  & \xmark\\
    \midrule
    \textbf{Misviz-synth (ours)} & 57,665 & 12 & 5 & 39 & \cmark & \xmark & \xmark & \cmark\\
    \textbf{Misviz (ours)}  & 2,604 &  12  & > 5  & 31 & \cmark & \cmark  &  \cmark  & \xmark\\
    
    \hline
  \end{tabular}}
  \caption{Existing datasets for misleading visualization detection.}
  \label{tab:literature-review}
\end{table*}

The first attempts to detect misleading visualizations relied on rule-based systems called linters \citep{mcnutt2018linting,10.1111/cgf.13975,9552878}. These linters assume the availability of the underlying data table or the chart code, which restricts their applicability to real-world scenarios. \citet{10.1145/3491102.3502138} and \citet{10.1145/3715336.3735784} overcome these limitations by extracting tables using OCR tools from real-world visualizations before applying rule checks. However, the accuracy of these real-world linters depends heavily on the quality of the intermediate OCR step, which can vary widely \citep{10.1145/3715336.3735784}. Real-world linters have previously been evaluated in small-scale user studies with human-in-the-loop correction of OCR errors.

Others have explored the potential of MLLMs for the task. \citet{10679256} evaluated four MLLMs with different prompts on a dataset of 150 real-world visualizations, sourced from the corpus of \citet{lo2022misinformed} for misleading cases. They found that detection accuracy decreased as more misleader types were included in the prompt. \citet{10771149} focused on GPT-4 \citep{openai2023gpt4}, using visualizations from the social media platform X \citep{10.1145/3544548.3580910}. However, access to this dataset requires a paid API, and reproducibility is further hindered by the platform's frequent removal of posts. In a parallel work, \citet{das2025misvisfixinteractivedashboarddetecting} proposed a prompt with which SOTA MLLMs achieve high accuracy on a subset of the corpus of \citet{lo2022misinformed}.

Recently, \citet{10.1007/978-3-031-98414-3_7} fine-tuned a convolutional neural network for the task, achieving high accuracy on synthetic instances.

Table~\ref{tab:literature-review} compares prior datasets with Misviz and Misviz-synth. Misviz is over fifteen times larger than the dataset of \citet{10679256}. Unlike \citet{10771149}, it does not rely on paid APIs for data collection and ensures long-term access to all instances by archiving them on the Wayback Machine.\footnote{\href{https://web.archive.org/}{web.archive.org}} Misviz-synth is two to seven times larger than other synthetic datasets, and includes several more misleaders and chart types \citep{mischaqa,10.1007/978-3-031-98414-3_7}. In contrast to other synthetic datasets,  Misviz-synth provides the underlying table, code, and axis metadata. The latter is necessary to fine-tune DePlot for axis extraction and answer our research questions.

\section{Misviz}

\subsection{Selected misleaders}

Misviz covers 12 types of misleaders, selected from the 74 categories defined in the taxonomy of \citet{lo2022misinformed}, based on four key criteria. First, we excluded misleaders that are rarely observed in real-world scenarios. To determine this, we used misleader frequency statistics from the corpus of \citet{lo2022misinformed} and discarded all categories with fewer than 15 instances. Second, we removed reasoning misleaders, i.e., misleaders that do not directly break chart design rules and are deceiving only in the context of a specific claim \citep{10.1145/3544548.3580910}. Third, we remove misleaders which confuse rather than deceive. As noted by \citet{lo2022misinformed}, the taxonomy includes both misleaders that distort the underlying data, the focus of this work, and others that may hinder readability or clarity without altering the interpretation of the data, such as \textit{missing titles} or \textit{overplotting}. Fourth, we excluded misleaders that require specific domain knowledge to be identified. For example, using red to represent Democrats and blue to represent Republicans in a chart violates \textit{color conventions}, but detecting this misleader requires familiarity with U.S. politics. Such misleaders require domain expertise that is beyond the reach of crowdworkers.

We define each selected misleader briefly below \citep{lo2022misinformed,10.1145/3544548.3581406,10670488}. They cover together 62.3\% of all instances from the real-world corpus of \citet{lo2022misinformed}. Each of them is represented with an example in Figure \ref{fig:misleaders} and in Appendix \ref{sec:solution}.

\textbf{Misrepresentation}: the value labels displayed do not match the sizes of their visual encodings; e.g., bars may be drawn disproportionately to their corresponding numerical values.

\textbf{3D}: the visualization includes 3D effects, distorting the size of visual encodings.

\textbf{Truncated axis}: an axis does not start from zero, thus exaggerating differences between values.

\textbf{Inappropriate use of pie chart}: a pie chart does not display data in a part-to-whole relationship. 

\textbf{Inconsistent binning size}: a variable, such as years or ages, is grouped in unevenly sized bins.

\textbf{Discretized continuous variable}: a continuous variable is cut into discrete categories, thus exaggerating the difference between boundary cases.

\textbf{Inconsistent tick intervals}: the ticks in one axis are evenly spaced, but their values are not, e.g., the tick values sequence is 10, 20, 40, 45.

\textbf{Dual axis}: there are two independent and parallel numerical axes with different scales.

\textbf{Inappropriate use of line chart}: a line chart is used in unusual ways, e.g., with categorical data.

\textbf{Inappropriate item order}: the tick labels of an axis are sorted in an unconventional way, e.g., dates are not shown chronologically.

\textbf{Inverted axis}: an axis is displayed in a direction opposite to conventions.

\textbf{Inappropriate axis range}: the axis range is either too broad or too narrow, minimizing or exaggerating the real trend.

\subsection{Data collection}

We obtain visualizations from three sources.

(a) We collect instances from the corpus of \citet{lo2022misinformed} that contain at least one of the 12 selected misleaders. We apply perceptual hashing to remove duplicates. We then manually discard instances with low resolution, those that display the name of the misleader, or those that show both a misleading chart and a corrected version side by side. Appendix~\ref{sec:discarded} shows removed examples.

(b) We use the misleading visualizations from the website WTF Visualizations,\footnote{\href{https://viz.wtf/}{viz.wtf}}  previously annotated by \citet{10670488} for taxonomy construction. We align their misleader categories with those of \citet{lo2022misinformed}, as explained in Appendix \ref{sec:wtfviz}, retain only those matching one of our 12 target misleaders, and remove duplicates.

(c) We access a large collection of unlabeled visualizations from \textit{r/dataisugly} and \textit{r/dataisbeautiful}.\footnote{\href{https://www.kaggle.com/datasets/bcruise/reddit-data-is-beautiful-and-ugly}{kaggle.com/datasets/bcruise/reddit-data-is-beautiful-and-ugly}} The former is an online community focused on sharing misleading visualizations, while the latter features high-quality examples and serves as our source of non-misleading instances. We begin by removing duplicates, then hire annotators to assign labels, as explained in Section \ref{sec:labeling}.

\subsection{Data labeling}
\label{sec:labeling}

We split the labeling of visualizations from \textit{r/dataisugly} and \textit{r/dataisbeautiful} into three annotation tasks. For each task, we hired three crowdworkers from Prolific and paid them £9 per hour. To ensure sufficient familiarity with visualizations, we required annotators to hold a PhD degree. Inter-annotator agreement (IAA) was evaluated using Fleiss'$\kappa$ \citep{fleiss1971measuring} on an overlapping set of 30 instances across all crowdworkers. Appendix \ref{sec:prolific} shows the annotation interface and instructions.

In the first task, crowdworkers assigned one or more chart types to each visualization. Five categories were available: bar charts, line charts, maps, scatterplots, and others. The ``others'' category includes less frequent chart types such as treemaps. The crowdworkers achieved a high IAA  (0.71).

In the second task, crowdworkers labeled visualizations from \textit{r/dataisugly} with zero to three misleaders. They received detailed guidelines with definitions and examples for each misleader. Some visualizations were assigned no label and removed from the corpus, typically because they contained a misleader outside of our selected set. A moderate IAA of 0.53 is achieved. This is a reasonable outcome, given the task’s complexity with 12 categories and multiple possible combinations.

In the final task, crowdworkers verified that visualizations from \textit{r/dataisbeautiful} were not misleading. Visualizations identified as containing one or more misleaders were removed from the corpus. The Fleiss'$\kappa$ was 0.78, reflecting high IAA.

\subsection{Bounding box labeling}

We recruited ten PhD students to annotate misleading visualizations using the VIA tool \citep{dutta2016via,dutta2019vgg}. Annotators drew bounding boxes around relevant misleader features, e.g., the initial tick mark on a truncated axis. Three misleaders were excluded: \textit{misrepresentation}, \textit{3D}, and \textit{inappropriate use of pie chart}, because bounding boxes are not suitable for representing them. Each student annotated up to 95 instances. We calculated IAA on a shared subset of 27 instances, three per misleader. Using an Intersection over Union (IoU) threshold of 0.4, the annotators achieved an IAA of 0.81, indicating strong agreement.

\subsection{Data statistics}

Misviz is divided into a few-shot development set (5\%), a validation set for hyperparameter and prompt tuning (15\%), and a held-out test set (80\%). 78\% of the visualizations contain one of the three main chart types: bar chart, line chart, or pie chart. While 94\% of the visualizations contain a single chart type, 5 and 1\% contain two or three different chart types, respectively. The split is stratified to ensure a balanced distribution of misleaders and chart types. Among the misleading visualizations,  85, 14, and 1\% contain one, two, or three misleaders, respectively.  The most frequent misleader is misrepresentation, present in 32\% of the visualizations. This aligns with prior findings \citep{lo2022misinformed,10670488}.  The next most frequent categories are 3D effects and truncated axes, appearing in 14 and 9\% of the visualizations, respectively. All other misleaders occur in 1 to 5\% of the instances. Appendix \ref{sec:stats} provides detailed distributions of misleaders and chart types. 

Although we do not provide language distributions, it is worth noting that the visualizations span many languages. The types of images also vary significantly. While most are screenshots, some are pictures of visualizations printed on paper or displayed on screens. Appendix \ref{sec:diverse_vis} provides cases of non-English and non-screenshot visualizations.

We did not leverage the large unlabeled corpus collected by \citet{lo2022misinformed}. However, this resource could be used in future work to scale the dataset, as discussed in Appendix \ref{sec:semigroundtruth}.

\section{Misviz-synth}

\subsection{Data collection}

We collect real-world data tables from Our World in Data, an open-access data platform that covers a wide range of domains, including health, education, and the economy.\footnote{\href{https://ourworldindata.org/}{ourworldindata.org}}

\subsection{Synthetic visualizations generation}

\begin{figure}
    \centering
    \includegraphics[width=\linewidth]{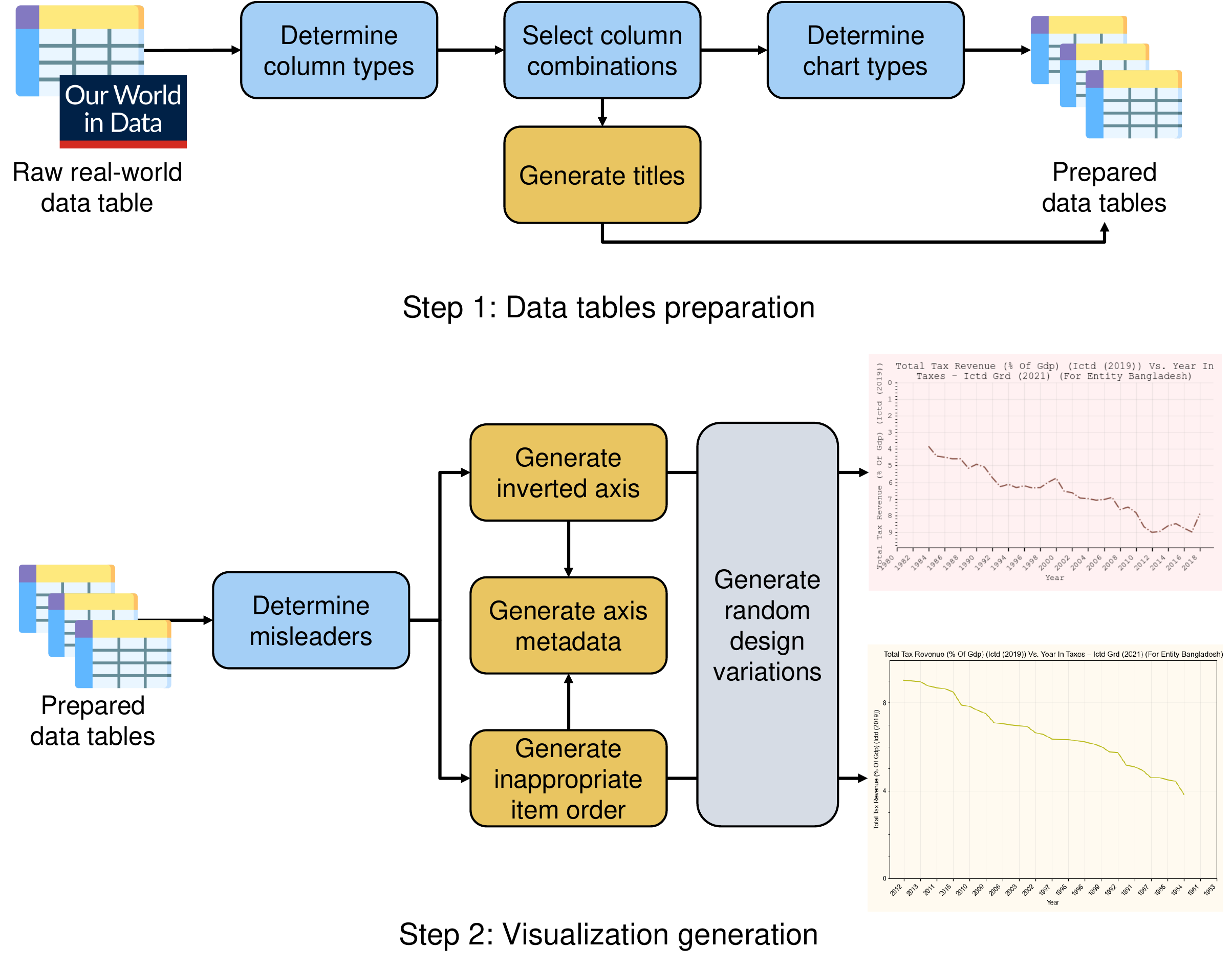}
    \caption{The two-step process to create the synthetic visualizations of Misviz-synth based on real-world data.}
    \label{fig:misviz-synth}
\end{figure}

We design a two-step rule-based system to generate synthetic visualizations, illustrated in Figure~\ref{fig:misviz-synth}.

In the first step, given a raw real-world data table, we assign one or more types to each column. There are 10 different column types, explained in Appendix \ref{sec:column_types}. Based on these types, we select valid column combinations that require at least one numerical column and a natural key for each row. This natural key can consist of one or more categorical or temporal columns. If the natural key involves multiple columns, we fix all but one to a constant value. For instance, if the numerical column is \textit{Win ratio} and the natural key includes \textit{Football club} and \textit{Year}, we set \textit{Year} to 2021 to compare the win ratios of football clubs for that year.  Chart titles are automatically generated using templates based on the selected columns. We then determine which chart types, among bar, line, and pie charts, are suitable for the data. For example, line charts are only compatible with temporal data, not categorical data. Each valid configuration produces a ``prepared'' data table.

In the second step, we generate visualizations using hand-crafted plotting functions for each chart type–misleader pair. For each prepared data table, we identify the misleaders applicable given the column types and selected chart types. One or more misleading visualizations are generated, each with exactly one misleader from the applicable set. Additionally, non-misleading visualizations are generated for a random subset of the prepared tables.

For bar, line, and scatter plots, we automatically extract axis metadata. Since pie charts and maps lack axes, they do not have axis metadata. Axis metadata is stored as a table with four columns. (1) \textit{Seq} indicates the sequential order of the tick marks along an axis. Horizontal axes are read from left to right, while vertical axes are read from bottom to top. (2) \textit{Axis} indicates the name of the axis, e.g., x, y1, or y2, in case of dual axis. (3) \textit{Label} is the tick label. (4) \textit{Relative position} is a normalized float indicating the tick mark's position, with spacing expressed relative to the distance between the first two tick marks. Examples of axis metadata are provided in Appendix~\ref{sec:axis_metadata}.

To increase visual diversity, we apply random variations in font size, background color, and axis-label positions. The complete list of variations is given in Appendix~\ref{sec:design_variations}. 

Each instance is saved along with its title, data table, Matplotlib code, and axis metadata.

\subsection{Data statistics}

Misviz-synth contains 57,665 visualizations, distributed into stratified train-large (80.1\%), train-small (9.6\%), dev (3.1\%), validation (3.2\%), and test sets (4.1\%). Misviz-synth covers the five most common chart types in Misviz: bar, line, and pie charts; scatterplots; and maps. Unlike Misviz, each visualization in Misviz-synth contains at most one chart type and one misleader, making it slightly less diverse than Misviz.

\section{Experiments}

\subsection{Baselines}

\begin{figure}
    \centering
    \includegraphics[width=\linewidth]{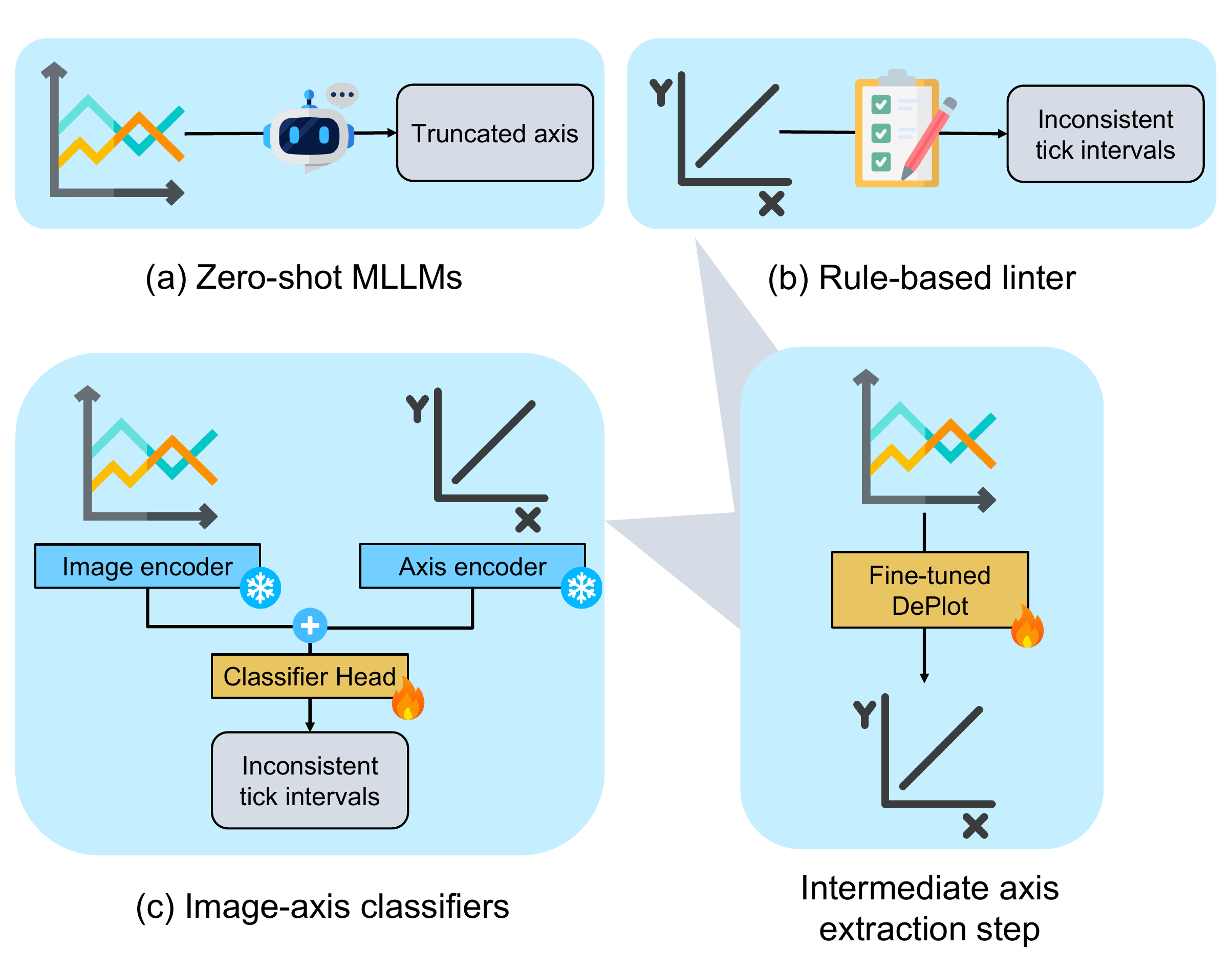}
    \caption{The three types of baselines included in the experiments. The linter and one classifier require axis extraction as an intermediate step.}
    \label{fig:baselines}
\end{figure}

We consider three categories of baseline models, illustrated in Figure \ref{fig:baselines}.

\textbf{Zero-shot MLLMs} We evaluate the zero-shot capabilities on one run of the following commercial and open-weight MLLMs: GPT-4.1 and GPT-o3 \citep{openai2023gpt4}, Gemini-2.5-Flash-Lite \citep{google2024gemini}, Qwen-2.5-VL in 7B, 32B, and 72B variants \citep{qwen2.5vl}, and InternVL3 in 8B, 38B, and 78B variants \citep{internvl3}. These models were selected for their strong performance on the ChartQA benchmark \citep{masry-etal-2022-chartqa}. The prompt includes the task description and definitions of the misleaders. The prompt is provided in Appendix~\ref{sec:prompt}.

\textbf{Rule-based linter} We design a linter that detects misleaders by applying rule-based checks to the axis metadata of a visualization. Each misleader corresponds to a specific rule. If a visualization passes all checks, it is classified as having no misleaders. We rely on axis metadata rather than the underlying data table because it enables detection of a broader range of misleaders: truncated axis, inverted axis, dual axis, inconsistent tick intervals, inconsistent binning size, and inappropriate item order. Only the latter two misleaders could also be detected from the data table alone. Detailed descriptions of all rule checks are provided in Appendix~\ref{sec:rules}. Misleaders that require visual interpretation or contextual knowledge, such as misrepresentation or inappropriate axis ranges, are not covered by the linter. We evaluate its performance using both ground truth and predicted axis metadata.

\textbf{Image-axis classifiers} We train two classifiers: one that takes only the visualization image as input, and another that combines the image with its axis metadata. Visualizations are encoded using a frozen image encoder, while axis metadata is embedded using a frozen table encoder. We use the image encoder of TinyChart, a specialized chart understanding MLLM \citep{zhang-etal-2024-tinychart}, while the table encoder is TaPas \citep{herzig-etal-2020-tapas}.   For the second classifier, the resulting image embeddings are concatenated with the [CLS] token from the axis metadata embeddings. The image embedding or concatenated embeddings are passed through a trained classification head. The classifiers are trained on Misviz-synth to predict either one misleader per visualization or none.

\textbf{Axis extraction} The linter and the second classifier require axis metadata as input. While ground truth metadata is available for Misviz-synth, this is not the case for Misviz or real-world scenarios, where only the visualization image is accessible. To address this, we implement an intermediate axis extraction step using DePlot \citep{liu-etal-2023-deplot}. Since DePlot was originally only trained for chart-to-table extraction, we fine-tune it on (image, axis metadata) pairs from Misviz-synth.

\subsection{Evaluation metrics}

We evaluate model performance using six metrics. The first four assess the binary classification of visualizations as misleading or not: Accuracy (Acc), Precision (Pre), and Recall (Rec) for the misleading class, and macro-F1 score (F1). The remaining two metrics are computed on the subset of visualizations that are misleading and evaluate the correct identification of specific misleaders. Exact Match (EM) assigns a score of 1 if the set of predicted misleaders exactly matches the ground truth. Partial Match (PM) assigns a score of 1 if the predicted misleaders are a subset of the ground truth.

\subsection{Implementation details}

Open-weight models are accessed via Transformers \citep{wolf-etal-2020-transformers}, while commercial models are accessed via their official APIs. For all MLLMs, we set the temperature to 0, except for GPT-o3, which uses the default value of 1. Results for the image-axis classifiers are reported as averages over three different seeds (123, 456, 789), with standard deviations reported in Appendix \ref{sec:std_classifier}. Additional information and hyperparameter values are provided in Appendix \ref{sec:details}.

\subsection{Results}

Table~\ref{tab:results_all_models} compares the performance of all models on the test sets of Misviz-synth and Misviz.

We organize our insights around the following RQs. \textbf{RQ1}: Which type of model performs best on real-world or synthetic instances? \textbf{RQ2}: Can detection models trained on synthetic instances generalize to real-world cases? \textbf{RQ3}: Can axis extraction models trained on synthetic instances generalize to real-world cases?

There are substantial performance differences across MLLMs, with GPT models achieving the best scores on both datasets, in particular for EM and PM. We attribute this to their strong OCR capabilities. For the Qwen2.5-VL family, performance improves with model scale. In particular, the largest version, 72B, is required to achieve EM scores above 11\%. For the InternVL3 family, the best model size depends on the dataset: 38B for Misviz, 78B for Misviz-synth.

MLLMs outperform linters and image-axis classifiers on the real-world visualizations of Misviz. This trend does not hold for Misviz-synth, where MLLMs are outperformed in both F1 and EM. We attribute this to two main factors. First, Misviz-synth instances often contain many axis ticks, making extracted axis metadata a particularly informative feature for misleader detection. Second, the axis extractor and image-axis classifiers directly benefit from being trained on the Misviz-synth train set, resulting in stronger in-domain performance.

\begin{tcolorbox}[colback=myblue!20, colframe=myblue!80, width=\linewidth]

\textbf{Insight 1.}  MLLMs perform better on Misviz, while the image-axis classifiers and linters are the best on Misviz-synth (RQ1).

\end{tcolorbox}

\begin{table*}
  
  \centering
  \resizebox{\textwidth}{!}{ %
  \begin{tabular}{lccccccccccccc}
    \hline    
       &  \multicolumn{6}{c}{Misviz} & & \multicolumn{6}{c}{Misviz-synth} \\
         \cline{2-7}  \cline{9-14} 
    
         & Acc & Pre & Rec & F1 &  EM & PM & & Acc & Pre & Rec & F1 &  EM & PM \\
    \hline
    & \multicolumn{13}{c}{\textit{Zero-shot MLLMs}} \\
    \hline
    Qwen2.5-VL-7B & 41.9 & 66.1 & 33.5 & 41.8 & 5.2 & 9.0 & & 54.5 & 68.7 & 52.9 & 52.8 & 10.8 & 10.8  \\
     Qwen2.5-VL-32B & 73.7 & 73.9 & 95.9 & 59.4 & 4.8 & 6.2 & &  65.3 & 66.8 & 93.5 & 48.7 & 3.7 & 3.7 \\
     Qwen2.5-VL-72B & 72.3 & 76.6 & 86.5 & 64.1 & 26.5 & 35.0 &  & 67.2 & 68.5 & 89.9 & 57.8 & 29.0 & 29.0 \\
    InternVL3-8B & 63.1 & 68.5 & 86.8 & 45.1 & 22.3 & 28.9 & & 63.1 & 65.4 & 82.7 & 44.2 &  10.2 & 10.2 \\
    InternVL3-38B & 59.2 & 76.5 & 59.4 & 56.8 & 28.5 & 37.3 & & 58.3 & 73.0 & 57.6 & 57.1 & 27.9 & 27.9\\
    InternVL3-78B & 55.6 & 68.1 & 67.8 & 48.0 & 23.8 & 30.9 & & 63.0 & 65.3 & 89.2 & 50.1 & 33.2 & 33.2  \\
    Gemini-2.5-Flash-Lite  & 54.0  & 63.3 & 68.4 & 44.7 & 29.7 & 39.4 & & 64.6 & 65.4 &  \digitbf{94.3} & 48.7 & 22.0 & 22.0 \\
    GPT-4.1& \digitbf{84.1} & 84.5 &  \digitbf{94.3}  & 79.6 & 53.6  & 64.0 & &  67.1 & 72.7 & 77.4 & 63.4 & 42.7 & 42.7 \\  
    GPT-o3 &  83.5 & \digitbf{86.6}  &  90.1 & \digitbf{80.0} &  \digitbf{58.8} &  \digitbf{67.5} & & 68.2 &  78.4 & 69.2 & 66.9 &  44.3 &  44.3 \\
    \hline
    & \multicolumn{13}{c}{\textit{Rule-based linter}} \\
    \hline
    Linter w. axis (ground truth)  &   - & -  &  - & - & - & - & & 69.4 &  99.7 & 52.2 & 69.4 & 51.4 &  51.4 \\ 
    Linter w. axis (DePlot\includegraphics[height=1em]{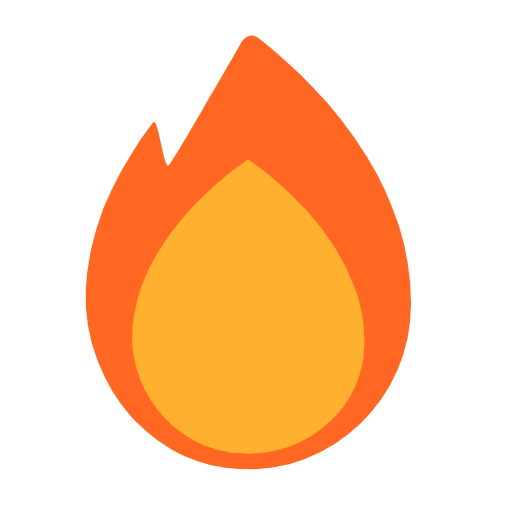}) & 36.5 & 63.1 & 20.4  & 36.1 & 6.6 & 7.8 & & 67.9 &  \digitbf{98.7} & 50.3 &  \digitbf{67.9} & 47.6  &  47.6\\
    \hline
    & \multicolumn{13}{c}{\textit{Image-axis classifiers}} \\
    \hline
    Image & 70.1  & 73.7  & 88.3  & 58.7 & 11.1 & 14.9  & & 72.0 & 71.9  & 92.2 & 64.7  & 68.0  & 68.0 \\
    Image w. axis (DePlot\includegraphics[height=1em]{figures/fire.png}) &  72.8 & 74.6 & 92.0 & 60.9  & 12.3  & 17.1  & & \digitbf{72.5}  & 72.1  & 92.6 & 65.2  &  \digitbf{69.5}  &  \digitbf{69.5}  \\
    \hline
    
  \end{tabular}}
  \caption{Performance (\%)  on the test sets of Misviz and Misviz-synth.  The best results, excluding the linter with ground truth axis metadata, are marked in \digitbf{bold}.}
  \label{tab:results_all_models}
\end{table*}

The linter with ground-truth axis metadata achieves high precision on Misviz-synth, demonstrating the usefulness of axis metadata for misleader detection and the precision guarantees offered by a rule-based system. However, the linter only covers six misleaders, negatively affecting recall and EM. Using predicted axis metadata rather than ground truth metadata has only a small negative effect on all metrics for Misviz-synth, with F1 and EM scores remaining above those of the best MLLMs. This indicates that the fine-tuned DePlot model reliably extracts axis information from Matplotlib-generated charts. However, on Misviz, the EM and PM are very low, dropping by around 40 points compared to Misviz-synth. This shows that DePlot fails to generalize to real-world visualizations because the diversity of axis designs in the Misviz-synth training instances is lower.

\begin{tcolorbox}[colback=myblue!20, colframe=myblue!80, width=\linewidth]

\textbf{Insight 2.}  The linter has high precision for a subset of misleaders, but is very sensitive to axis extraction errors (RQs 1 and 3).

\end{tcolorbox}

\begin{tcolorbox}[colback=myblue!20, colframe=myblue!80, width=\linewidth]

\textbf{Insight 3.}  The axis extractor fine-tuned on synthetic visualizations has limited generalizability to real-world ones (RQ 3).

\end{tcolorbox}

On Misviz-synth, the image-axis classifiers achieve higher EM and PM than all other baselines. The classifier with axis metadata is the strongest, further highlighting the usefulness of that input modality. Their scores for binary classification on Misviz decrease only slightly compared to Misviz-synth. As a result, the classifier with axis metadata outperforms all open-weight MLLMs except Qwen2.5-VL-72B. However, the EM and PM scores drop by more than 50 points from Misviz-synth to Misviz. This implies that the classifiers cannot generalize their ability to make fine-grained predictions of misleader categories from synthetic to real-world visualizations.

\begin{tcolorbox}[colback=myblue!20, colframe=myblue!80, width=\linewidth]

\textbf{Insight 4.}  Classifiers trained on Misviz-synth can generalize to Misviz for binary classification (RQ 2).

\end{tcolorbox}

The best EM scores remain low on both datasets, underscoring the difficulty of identifying which misleaders affect a visualization. For binary classification, most baselines tend to favor recall over precision, with the notable exception of the linter. Future work could aim for a better balance to improve overall F1 performance.

Based on these observations, rule-based linters and image-axis classifiers are best suited for controlled environments where table and axis metadata are available. They can assist chart designers by automatically detecting misleaders introduced unintentionally. In contrast, for flagging misleading visualizations encountered online, where only the image is available, MLLMs are the better option.

Appendix \ref{sec:misviz_per_source} reports the image-axis classifiers and linter results on Misviz per number of chart type and per number of misleader, providing deeper insights into the generalization gap. Appendix \ref{sec:sensitivity_analysis} provides a prompt sensitivity analysis using the best open-weight models.

\subsection{Manual error analysis}

\begin{table}
    \centering
    \resizebox{\linewidth}{!}{ %
    \begin{tabular}{lll}
    \hline
    
    & Misviz & Misviz-synth \\
    \hline
    Incorrect use of misleader definition & 14 & 13 \\
    Incorrect measurement of object size & 10 & 3 \\
    Incorrect extraction of axis metadata & 2 &  14 \\
    Complex chart type & 2 & 0 \\
    No detection of 3D effects & 2 & 0 \\
    \hline
    \end{tabular}}
    \caption{Manual analysis of 30 prediction errors with GPT-4.1 on Misviz and Misviz-synth test sets.}
    \label{tab:error_analysis}
\end{table}

We manually analyze two random samples of 30 incorrect predictions by GPT-4.1. Errors are categorized into different types, as reported in Table~\ref{tab:error_analysis}, with examples in Appendix~\ref{sec:error}.

Around half the errors in both datasets are due to incorrect uses of misleader definitions. For instance, several instances are incorrectly classified as \textit{dual axis} because they display multiple lines. However, this misleader applies only when there are two distinct vertical axes. On Misviz, several instances are misclassified as \textit{inappropriate use of pie chart} because the values shown do not sum to 100. GPT-4.1 fails to see that the pie slice labels are absolute values rather than percentages.

Many errors in Misviz are due to incorrect measurement of the sizes of visual encodings, such as pie slices and bar heights, which are critical for detecting \textit{misrepresentation} by comparing the measured sizes with the value labels.

In Misviz-synth, many errors arise from incomplete or incorrect parsing of axis metadata. For example, GPT-4.1 frequently overlooks \textit{inconsistent tick intervals} on long temporal or numerical axes with numerous tick marks.

Two additional error categories are unique to Misviz. Some real-world visualizations are complex in their structure, sometimes containing more than four charts, which overwhelms the model with excessive visual information. In other cases, GPT-4.1 fails to detect \textit{3D} effects, the only misleader that does not require analyzing any displayed labels.

\subsection{Bounding box generation}

\begin{figure}
    \centering
    \includegraphics[width=\linewidth]{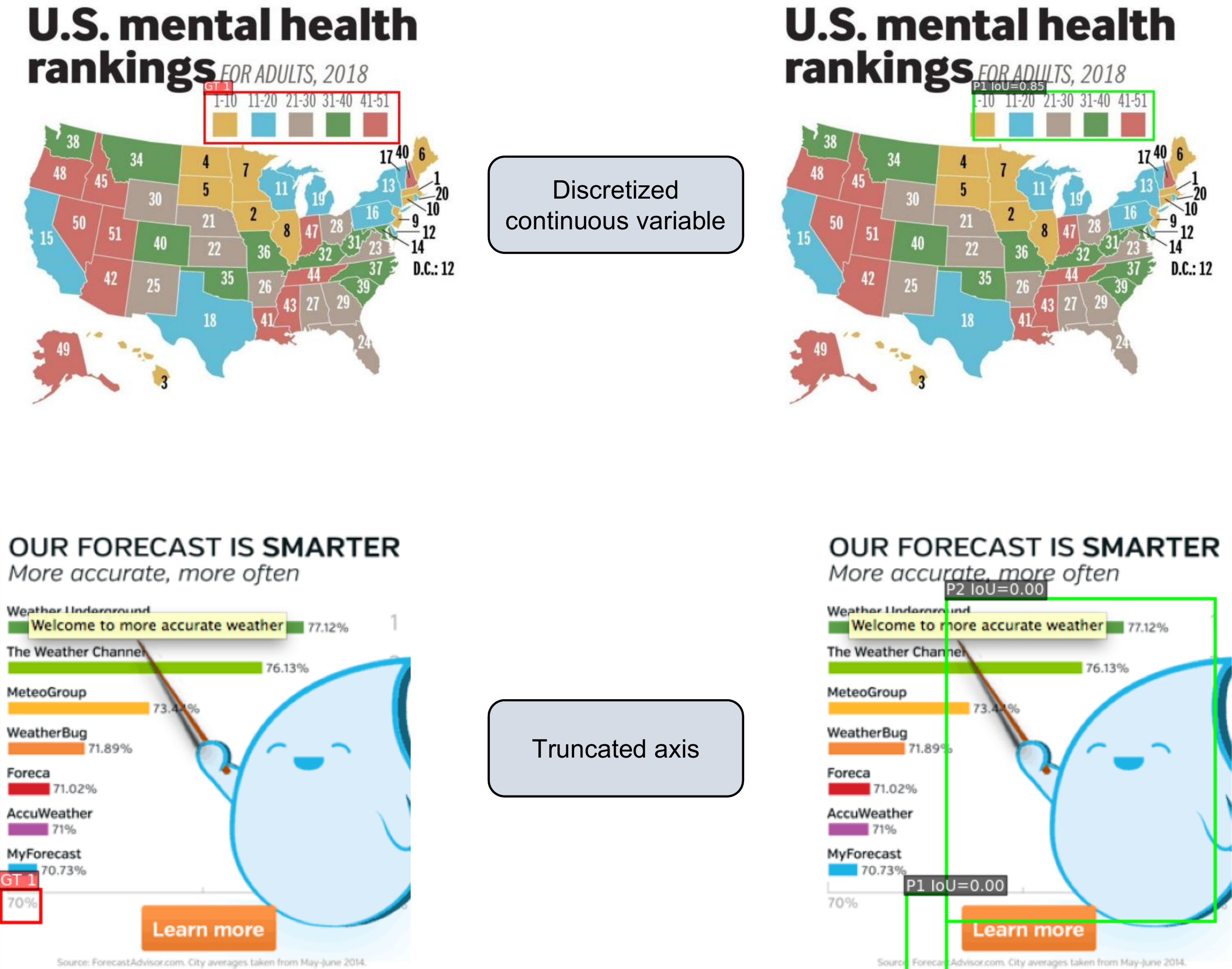}
    \caption{Examples of bounding box predictions for two instances of Misviz. The ground truth boxes are shown on the left, and the predictions on the right.}
    \label{fig:bbox}
\end{figure}

We conduct a small-scale experiment to assess the ability of MLLMs to localize misleaders in visualizations using bounding boxes. Such localized predictions can serve as a complementary form of explanation for end users, alongside the predicted misleader labels. We evaluate Qwen2.5-VL-72B, the strongest open-weight MLLM, on the Misviz val set. The prompt is provided in Appendix \ref{sec:prompt}. We use IoU as a metric, matching each predicted bounding box to its nearest ground-truth. The MLLM achieves an IoU of 13.1\%, a low score indicating that localization is challenging. Figure \ref{fig:bbox} shows two examples. In the top example, the MLLM accurately predicts a bounding box around the legend, indicating the discretized scale. In the bottom example, the MLLM fails to select the starting tick on the horizontal axis, indicating truncation.

\section{Conclusion}

In this work, we introduce two datasets for misleading visualization detection, covering 12 types of misleaders. Misviz is a real-world benchmark. Misviz-synth is a synthetic dataset generated from real-world tables, suitable for both training and evaluation. We propose a new linter and a trained classifier that leverage the axis metadata extracted from a visualization by a fine-tuned DePlot model. We conduct a comprehensive evaluation with three types of models: zero-shot MLLMs, linters, and image-axis classifiers.   Our results show that the task poses several challenges. MLLMs perform best on real-world visualizations, while image-axis classifiers have an edge on synthetic ones.

\section*{Limitations}

We identify three main limitations in this work.

First, Misviz-synth is less diverse than its real-world counterpart, Misviz. It does not include visualizations with multiple charts or multiple misleaders. Additionally, using the Matplotlib library imposes constraints; for instance, it does not support 3D pie charts. Future work should explore alternative plotting libraries to broaden coverage across chart types and misleader categories, thereby improving generalization to Misviz.

Second, the set of 12 misleader categories addressed in this work represents only a subset of those found in practice. While our selection was carefully made based on three criteria, future efforts should expand the taxonomy to include more misleaders. Notably, it would be valuable to incorporate misleaders that require domain knowledge to detect, such as \textit{violating color conventions}. Furthermore, our selection focuses on design misleaders. However, there are also reasoning misleaders in which no design rules are broken. Instead, they mislead through incomplete and dubious data \citep{lo2022misinformed}, or deceptive titles and annotations \citep{10.1145/3544548.3580910,10771149}. Such reasoning misleaders would require very different detection approaches and deserve more focus in future work.

Third, the boundaries between some misleader categories are not always clear-cut. For instance, \textit{inappropriate item order} and \textit{inverted axis} could be considered equivalent when a temporal axis is shown in reverse chronological order. Most cases of \textit{inappropriate item order} with maps are usually also cases of \textit{discretized continuous variables}, although only the former appears as the dominant misleader label in Misviz-synth. These label ambiguities are not captured by the current EM metric, which may slightly underestimate models' true performance in detecting misleading visualizations.

\section*{Ethics statement}

\textbf{Social impact} Misleading visualizations are frequently used by malicious actors to spread misinformation online, particularly on social media platforms \citep{2017-blackhatvis,10.1145/3544548.3580910}. They may also result from unintentional design choices by chart designers. This work contributes both a dataset and new baselines to help mitigate the negative impact of misleading visualizations by enabling their automatic detection. The Misviz and Misviz-synth datasets are intended solely for academic research. As with other resources developed for misinformation detection, there is a potential risk of dual use, where malicious actors exploit detection models in adversarial settings to craft misleading content that evades detection. However, we believe that the potential benefits of this research, such as assisting chart designers and protecting readers, outweigh these risks.

\textbf{Misinformation content} Misviz includes real-world disinformation examples. To preserve the authenticity and diversity of misleading visualizations encountered in the wild, we did not filter or censor such content.

\textbf{Dataset access} Our code and dataset annotations are released under the Apache 2.0 and CC BY-SA 4.0 licenses, respectively. We do not hold the rights to the visualization images in Misviz. Therefore, we do not distribute them directly. Instead, we provide image URLs in the Misviz dataset file, along with a script to download them. To ensure long-term accessibility, we have verified that all images are available on the WaybackMachine and included the corresponding archive URLS in the dataset file.

\textbf{AI assistants use}  AI assistants were used in this work to assist with writing by correcting grammar mistakes and typos.

\section*{Acknowledgments}

This work has been funded by the LOEWE initiative (Hesse, Germany) within the emergenCITY center (Grant Number: LOEWE/1/12/519/03/05.001(0016)/72), by the German Federal Ministry of Research, Technology and Space and the Hessian Ministry of Higher Education, Research, Science and the Arts within their joint support of the National Research Center for Applied Cybersecurity ATHENE, and by the Flanders AI Research Program. Figures \ref{fig:misviz-synth}, \ref{fig:baselines}, \ref{fig:error_misviz}, and \ref{fig:error_misviz_synth} have been designed using resources from Flaticon.com.  We thank Liesbeth Allein, Luke Bates, Manisha Venkat, Max Glockner, and Vivek Gupta for our insightful discussions on misleading visualizations. We are grateful to Niklas Traser for preparing the annotation interface for crowdworkers and implementing parts of the code for generating Misviz-synth. We also thank Shivam Sharma, Shivam Sharma, and Germán Ortiz for their feedback on an early draft of this work.

\bibliography{anthology,custom}

\appendix

\begin{figure*}[!ht]
    \centering
    \includegraphics[width=\linewidth]{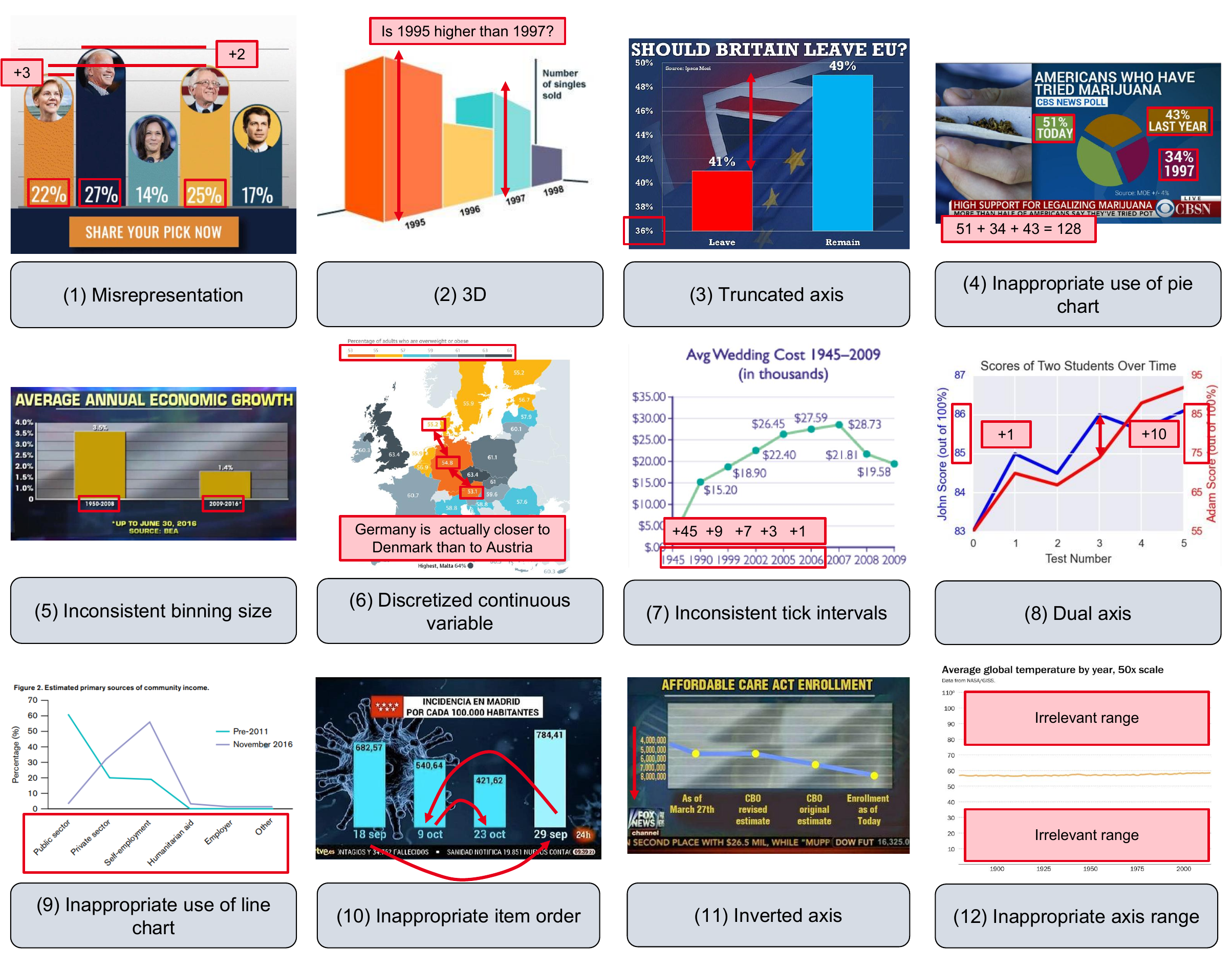}
    \caption{Examples of misleading visualizations from Figure \ref{fig:misleaders}, overlaid with visual explanations.}
    \label{fig:solution}
\end{figure*}

\section{Explanation of Misviz examples}
\label{sec:solution}

Figure~\ref{fig:solution} shows the same misleading visualizations as in Figure~\ref{fig:misleaders}, now overlaid with visual explanations highlighting the specific misleaders. We describe each case below:

\textbf{(1) Misrepresentation}  In terms of percentages, Bernie Sanders is closer to Joe Biden than to Elizabeth Warren. However, the fourth bar (Sanders) is visually closer in height to the first (Warren) than the second (Biden), exaggerating the gap between Biden and Sanders.

\textbf{(2) 3D}  The 3D effects make it difficult to compare values across years visually. For instance, it's unclear which year had the highest number of singles sold between 1995 and 1997.

\textbf{(3) Truncated axis} The  vertical axis begins at 36\%, exaggerating the gap between pro- and anti-Brexit responses.

\textbf{(4) Inappropriate use of pie chart} The pie chart shows responses to three overlapping time categories: today, last year, and 1997. Because respondents may fall into more than one category, the percentages exceed 100\%. The pie chart suggests the categories are distinct.

\textbf{(5) Inconsistent binning size} The two bars compare economic growth over different time spans. The first covers a longer period than the second, making the comparison misleading.

\textbf{(6) Discretized continuous variable} Germany and Denmark have similar values, but are shown in different colors. Meanwhile, Austria shares Germany’s color despite being further apart in value. The discrete color scale exaggerates differences between countries.

\textbf{(7) Inconsistent tick intervals} The dates on the horizontal axis are spaced unevenly. This distorts the slope of the trend line and gives a misleading impression of the progression of average wedding costs over time.

\textbf{(8) Dual axis} The scores for John and Adam are plotted on separate vertical axes. While the visual gap on test 3 looks small on the left axis, the actual difference is 11\%, not 1\%.

\textbf{(9) Inappropriate use of line chart} A line chart is used to connect values across a categorical variable. The line implies a trend that does not exist.

\textbf{(10) Inappropriate item order} The dates are out of chronological order, which can mislead to thinking COVID-19 cases initially dropped and then rose. In fact, the reverse is true.

\textbf{(11) Inverted axis} The vertical axis increases from top to bottom. At first glance, it seems that enrollment in the Affordable Care Act is falling over time, while it is actually rising.

\textbf{(12) Inappropriate axis range} The vertical axis is so broad that the trend in global average temperature appears nearly flat. A more appropriate narrow axis range would reveal a clearer upward trend.

\section{Example of discarded visualizations}
\label{sec:discarded}

\begin{figure}
    \centering
    \includegraphics[width=0.9\linewidth]{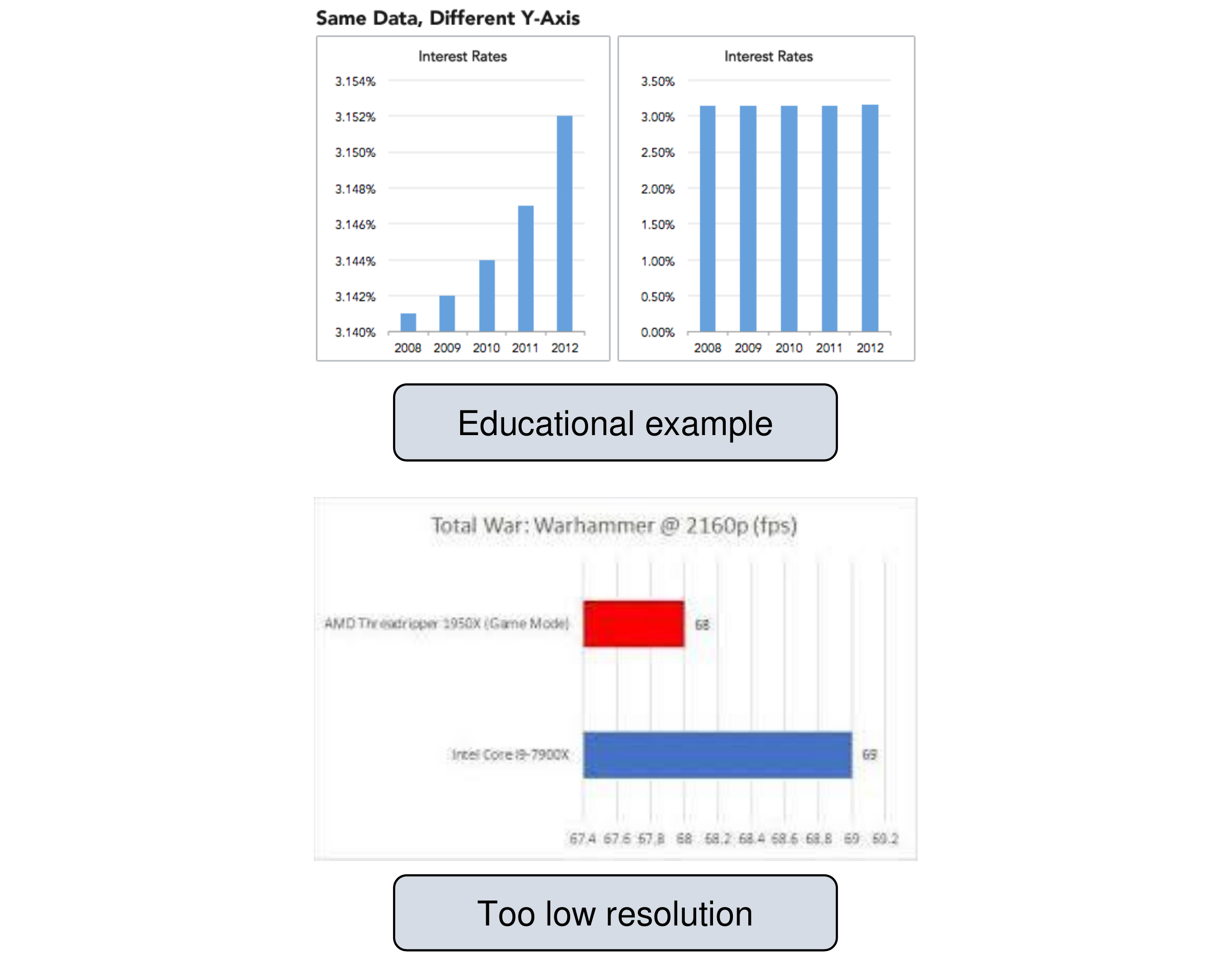}
    \caption{Examples of visualizations discarded from the corpus of \citet{lo2022misinformed}. Top: The misleading visualization is shown alongside a corrected version. Bottom: The visualization is too low-resolution.
}
    \label{fig:removed}
\end{figure}

\begin{table}
    \centering
    \resizebox{\linewidth}{!}{ %
    \begin{tabular}{ll}
    \hline
    \citet{10670488}  & \citet{lo2022misinformed}\\
    \hline
      Data-visual disproportion & Misrepresentation    \\
      3D effect & 3D \\
      Truncated axis & Truncated axis \\
      Misuse of part-to-whole relationship (pie chart) & Inappropriate use of pie chart \\
      Uneven axis interval & Inconsistent tick intervals \\
      Dual axis & Dual axis \\
      Uneven data grouping & Inconsistent binning size \\
      Categorical encoding for continuous data & Discretized continuous variable \\
      Continuous encoding for categorical data (line chart) & Inappropriate use of line chart \\
      Inverted axis & Inverted axis \\
      \hline
    \end{tabular}}
    \caption{Mapping of misleaders from the taxonomy of \citet{10670488} to the taxonomy of \citet{lo2022misinformed}.}
    \label{tab:wtfviz}
\end{table}

Figure \ref{fig:removed} provides two examples from the corpus of \citet{lo2022misinformed} that were not included in Misviz.

\section{Mapping of misleader categories between taxonomies}
\label{sec:wtfviz}

Table~\ref{tab:wtfviz} presents the manual mapping applied between the misleader categories from the taxonomy of \citet{10670488} to those of \citet{lo2022misinformed}.

\section{Annotation interface and instructions}
\label{sec:prolific}

\begin{figure*}
    \centering
    \includegraphics[width=\linewidth]{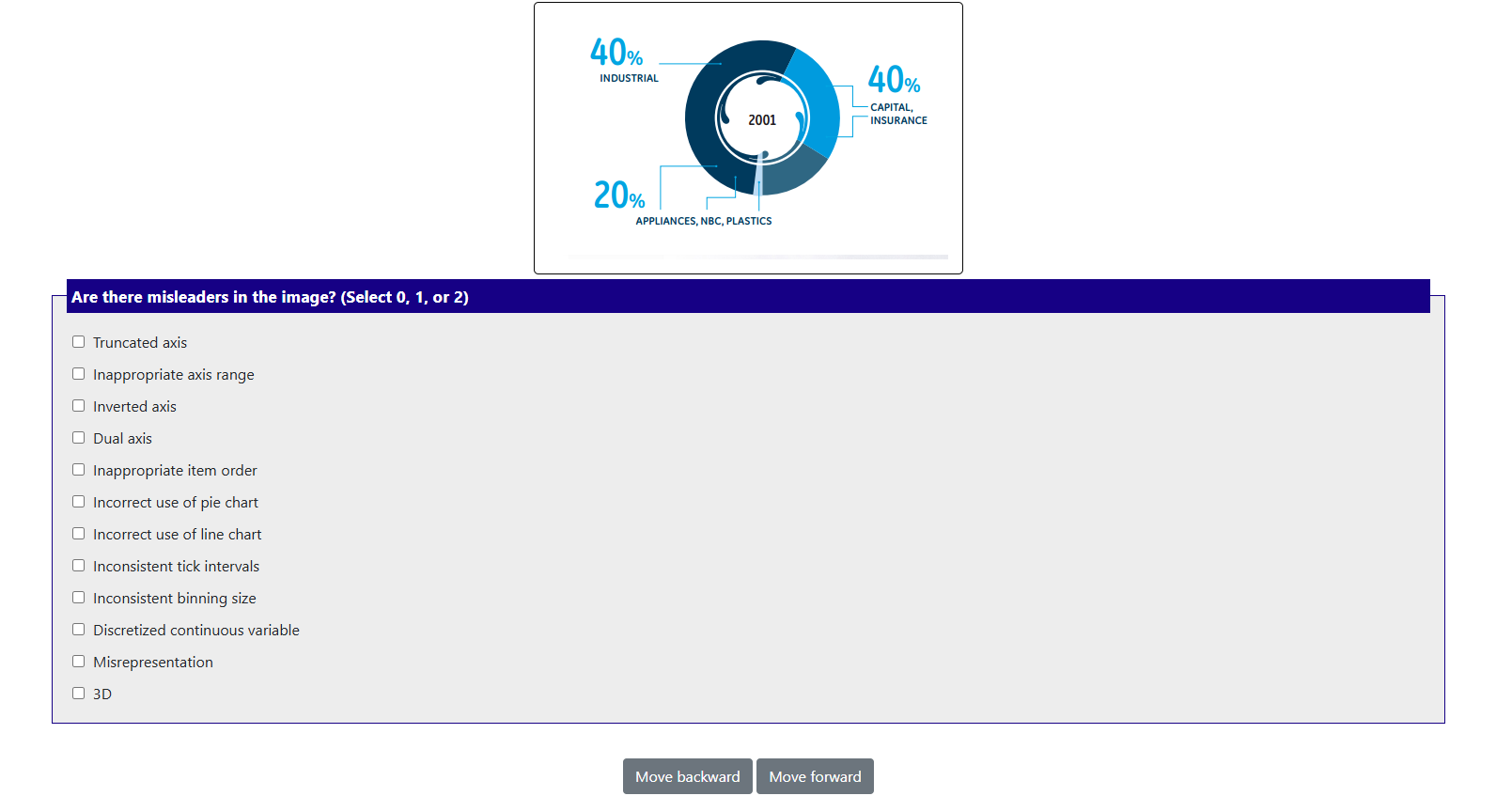}
    \caption{Main interface used by Prolific crowdworkers to assign misleader labels to a visualization.}
    \label{fig:prolific}
\end{figure*}

\begin{figure}[!ht]
    \centering
\begin{tcolorbox}[colback=gray!5, colframe=gray!80, title=Task 1: Chart types, width=\linewidth]

 In this study, you will be asked to classify a chart into the correct categories (pie chart, line chart, bar chart, map, scatter plot, or other). 
You are expected to spend, on average, 10 to 15 seconds on each chart.
 \\

 Select one or more types of charts that appear in the image. You should always select at least one. You can choose multiple answers if the image contains more than one type.
\end{tcolorbox}

    \caption{Instructions for the first annotation task.}
\label{fig:task1}
\end{figure}

\begin{figure*}[!ht]
    \centering
\begin{tcolorbox}[colback=gray!5, colframe=gray!80, title=Task 2: Misleading visualizations, width=\linewidth]

In this study, your task is to detect misleaders present in a chart. Misleaders are design flaws that decrease the ability of readers to interpret charts correctly. You are expected to spend, on average, 30 seconds to 1 minute on each chart. \\

Misleaders are chart design elements that can lead readers to interpret a wrong message that does not correspond to the underlying data. 
This study covers 12 types of misleaders.
Each chart contains one to three misleaders (one being the most frequent case). Your task is to select the correct misleader(s) for each chart.
We present the types of misleaders below. \\

 Important remarks \\
 - To perform this task, you need to pay attention to all aspects of the chart: axes and their labels, titles, legends, and proportions. Misleaders often reside in small details.\\
-It is essential to select the correct misleader(s) that apply to the image\\
Most charts contain only one misleader. In some cases, there will be two or three. \\

\end{tcolorbox}

    \caption{Instructions for the second annotation task.}
\label{fig:task2}
\end{figure*}

\begin{figure*}
    \centering
\begin{tcolorbox}[colback=gray!5, colframe=gray!80, title=Task 3: Non-misleading visualizations, width=\linewidth]

 In this study, your task is to verify that a chart is not misleading, i.e., it does not have significant design flaws that decrease the ability of readers to interpret charts correctly. Your participation will help us build AI systems capable of classifying charts as misleading or not. 
You are expected to spend, on average, 30 seconds to 1 minute on each chart. 
 \\

You will have to review charts that come from an online community. The majority of the charts are expected to be of high quality. However, some of them might be misleading. Misleaders are chart design elements that can lead readers to interpret a wrong message that does not correspond to the underlying data. There are 12 types of misleaders to consider, and they are explained below.
If the chart is not misleading (i.e., in MOST CASES), you should leave the answer field blank. If the chart is misleading, select the misleaders that apply (up to three misleaders).\\

Important remark\\
- To perform this task, you need to pay attention to all aspects of the chart: axes and their labels, titles, legends, and proportions. Misleaders often reside in small details.\\
- You should not select a misleader for each chart! Most of them are not misleading. You should only detect the few that are misleading!

\end{tcolorbox}

    \caption{Instructions for the third annotation task.}
\label{fig:task3}
\end{figure*}

\begin{figure}[!ht]
    \centering
    \includegraphics[width=\linewidth]{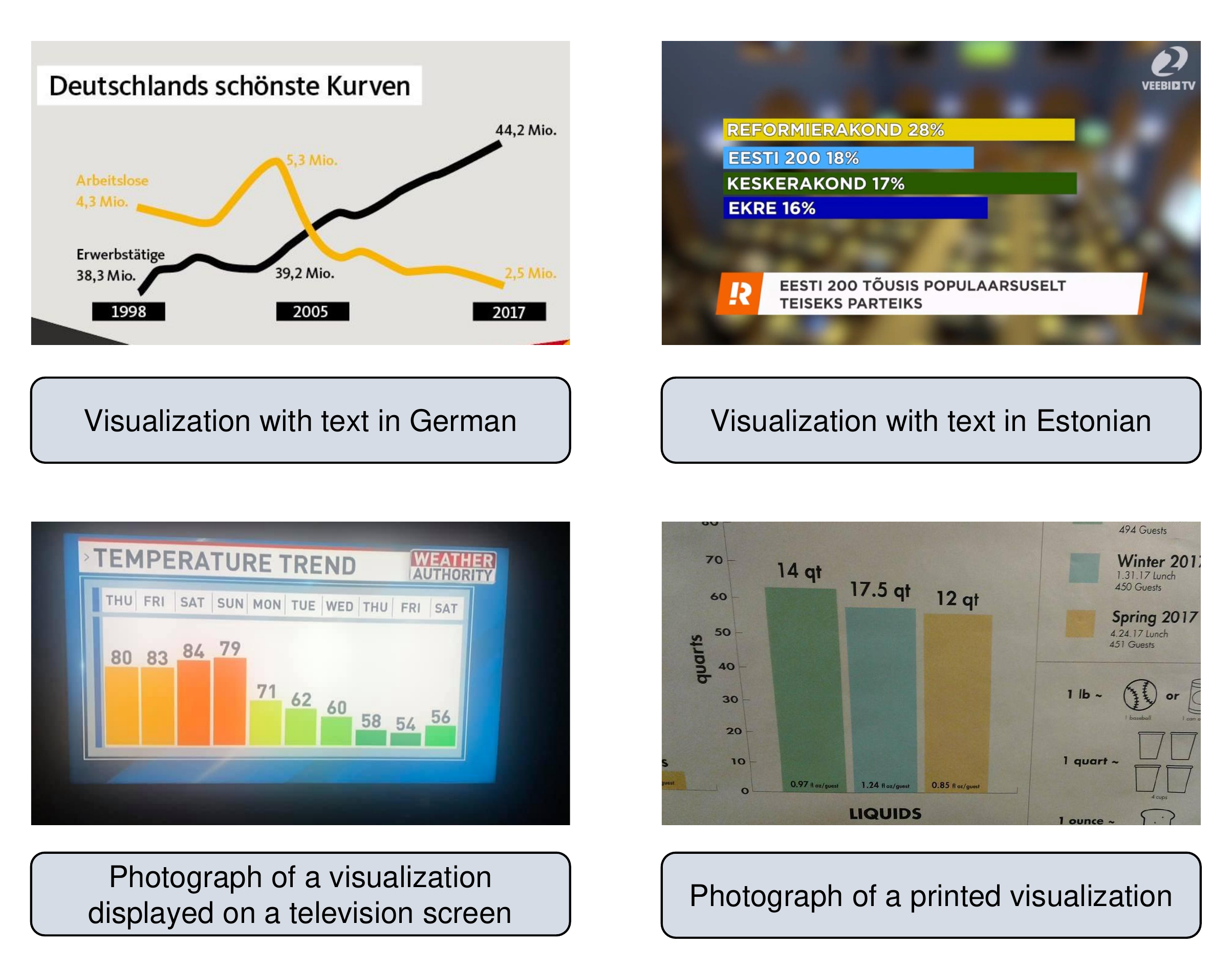}
    \caption{Four visualizations from Misviz. Top: the embedded text is not written in English. Bottom: the visualization is not a screenshot but a photograph taken of an electronic device or printed material.}
    \label{fig:diversity}
\end{figure}

\begin{figure}[!ht]
    \centering
    \includegraphics[width=\columnwidth]{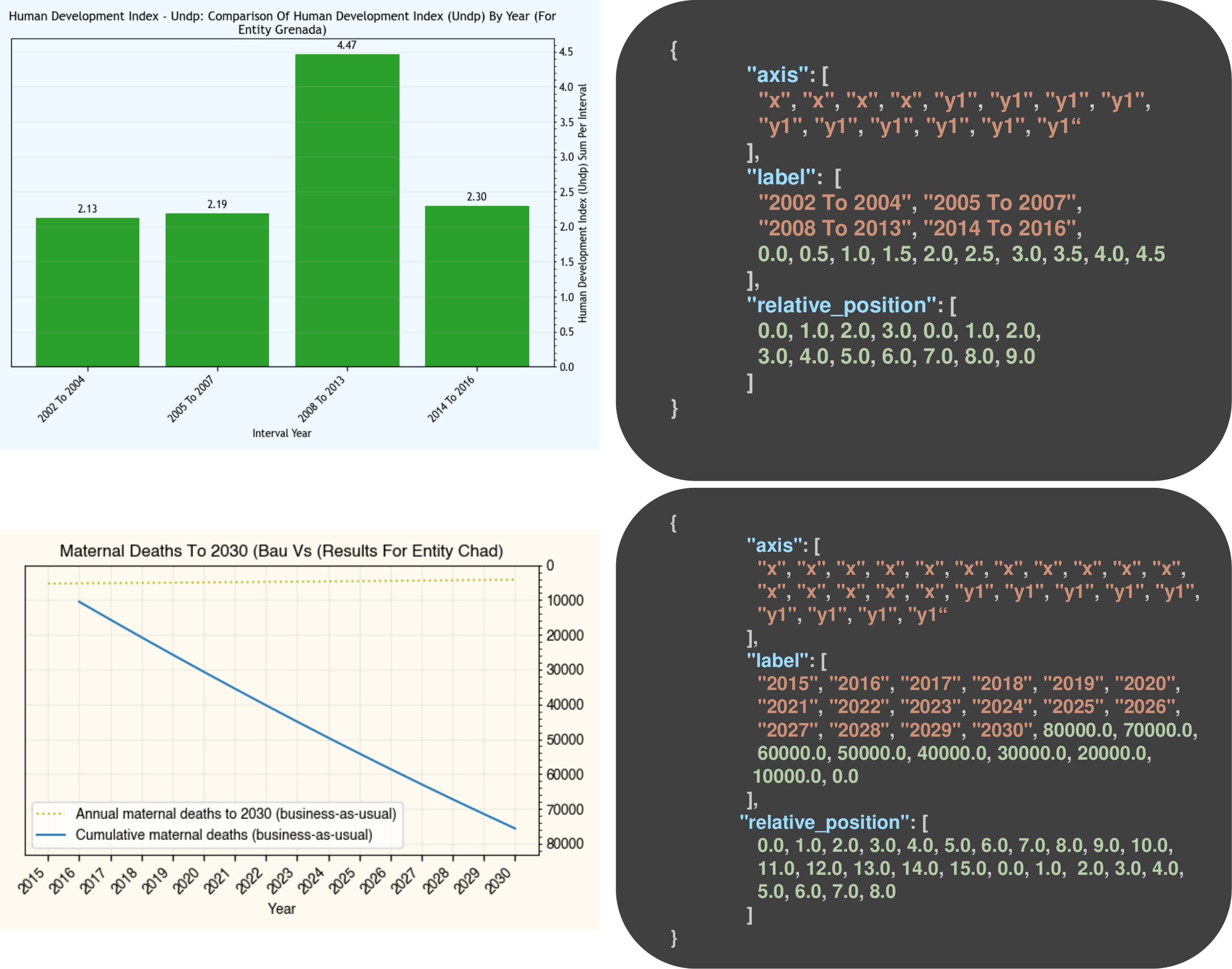}
    \caption{Two visualizations from Misviz-synth with their corresponding axis metadata.}
    \label{fig:axis_metadata}
\end{figure}

Figure \ref{fig:prolific} shows the main interface used by Prolific crowdworkers to label the visualizations from \textit{r/dataisugly} and \textit{r/dataisbeautiful}.

Figures \ref{fig:task1}, \ref{fig:task2}, and \ref{fig:task3} show the labeling instructions given to the crowdworkers.

\section{Detailed dataset statistics}
\label{sec:stats}

Tables \ref{tab:misleader_dist} and \ref{tab:chart_dist} report the distributions of misleader categories and chart types in Misviz and Misviz-synth. The percentages do not sum to 100\% in Misviz, because a visualization may contain more than one misleader category or chart type.

Of the 1,791 misleading visualizations in Misviz, 353 originate from the corpus of \citet{lo2022misinformed}, 964 were collected from the WTF Visualizations website \citep{10670488}, and the remaining 474 were sourced from \textit{r/dataisugly} and annotated by crowdworkers.

\begin{table}
    \centering
    \resizebox{\linewidth}{!}{ %
    \begin{tabular}{lll}
    \hline
    Misleader category     &  Misviz & Misviz-synth\\
    \hline
      Misrepresentation   & 32.6 & 5.5  \\
      3D & 14.0 & 2.2 \\
      Truncated axis &  8.7 & 0.8 \\
      Inappropriate use of pie chart & 6.2 & 1.2 \\
      Inconsistent tick intervals & 5.1 & 15.8 \\
      Dual axis & 3.0 &  3.0\\
      Inconsistent binning size & 2.8 &  2.1 \\
      Discretized continuous variable & 1.9 & 4.0 \\
      Inappropriate use of line chart & 1.8 &  2.2 \\
      Inappropriate item order & 1.5 & 8.4 \\
      Inverted axis & 1.2 & 11.9 \\
      Inappropriate axis range & 1.1 & 4.0\\
    \hline
    \end{tabular}}
    \caption{Percentage of instances per misleader.}
    \label{tab:misleader_dist}
\end{table}

\begin{table}
    \centering
    \resizebox{0.8\linewidth}{!}{ %
    \begin{tabular}{lll}
    \hline
    Chart type     &  Misviz & Misviz-synth\\
    \hline
      Bar   & 37.0 &  21.6 \\
     Line   & 20.0 & 33.5 \\
      Pie  & 23.4 & 3.6 \\
    Map   & 4.9  &  16.9\\
    Scatter plot & 4.3 & 24.4\\
    Other   & 16.5  &  0.0\\
    \hline
    \end{tabular}}
    \caption{Percentage of instances per chart type.}
    \label{tab:chart_dist}
\end{table}

\section{Examples of language and image type diversity in Misviz}
\label{sec:diverse_vis}

Figure \ref{fig:diversity} shows four instances highlighting the diversity of Misviz. The two visualizations at the top have embedded text written in German and Estonian. The bottom visualizations are photographs of a visualization shown on a television screen and on a printed document.

\section{Extending Misviz with weakly labeled corpus}
\label{sec:semigroundtruth}

\begin{figure}
    \centering
\begin{tcolorbox}[colback=gray!5, colframe=gray!80, title=Prompt to detect visualizations, width=\linewidth]

You are an expert in data visualization analysis. 
Your task is to identify whether the following image shows a data visualization or not.
Provide only the final answer (visualization OR not a visualization), without additional explanation.

\end{tcolorbox}
    \caption{Prompt used to detect whether an image shows a visualization.}
\label{fig:is_vis_prompt}
\end{figure}

\begin{table}
    \centering
    \begin{tabular}{ll}
    \hline
    Label & Precision \\
    \hline
    No misleader & 80 \\
    Misrepresentation & 40 \\
    3D & 100 \\
    Truncated axis & 20 \\
    Inappropriate use of pie chart & 60\\
    Inconsistent binning size & 40\\
    Discretized continuous variable & 70\\
    Inconsistent tick intervals & 90\\
    Dual axis & 30 \\
    Inappropriate use of line chart & 0 \\
    Inappropriate item order & 20 \\
    Inverted axis & 20\\
    Inappropriate axis range & 30 \\

      \hline
    \end{tabular}
    \caption{Manual verification of the weakly labeled corpus annotated with Qwen2.5VL-72B (\%).}
    \label{tab:manual_check_corpus}
\end{table}

\citet{lo2022misinformed} scraped a large collection of images to construct their taxonomy of visualization misleaders. The images obtained via web searches for keywords related to misleading visualizations. Only a small subset of this collection was manually annotated. In this section, we explore the potential of automatically scaling the dataset using this unlabeled collection.

We use a three-step process to create a weakly labeled corpus from the remaining unlabeled portion of the collection. First, we filter unreadable images with a heuristic: we extract texts from the image using EasyOCR\footnote{\href{https://github.com/JaidedAI/EasyOCR}{github.com/JaidedAI/EasyOCR}} and only keep images with an average OCR confidence over all detected text that is above 70. For the second and third steps, we use Qwen2.5VL-72B, the best performing open-weight model in terms of F1. In the second step, we ask the MLLM to determine whether the image depicts a visualization. The prompt is provided in Figure \ref{fig:is_vis_prompt}. Afterwards, we remove all images that do not depict a visualization. In the third step, we use the standard task prompt to detect which misleaders affect the visualization, if any, shown in Appendix \ref{sec:prompt}. We only keep images for which zero or one misleader is identified.

The resulting weakly labeled corpus contains 4,053 instances. We conducted manual verification of 10 instances per label, for a total of 130. As shown in Table \ref{tab:manual_check_corpus}, the results are uneven. For some misleaders, such as \textit{3D} and \textit{Inconsistent tick intervals}, Qwen2.5VL-72B achieves high precision on the manually verified sample, indicating that the corresponding subset of the corpus could be used to further train models to detect real-world cases of those misleaders. However, for other misleaders, such as \textit{inappropriate use of line chart}, the precision is so low that the weak labels should not be considered reliable. Furthermore, we observed that several unreadable images remained in the corpus despite the OCR filtering step. In addition, some images are educational examples that explicitly state they show a misleading visualization, as explained in Appendix \ref{sec:discarded}. Removing these examples would require an additional step of human verification. Overall, the weakly labeled corpus cannot be considered a reliable dataset, except for a few misleaders. Future research should further explore how to scale the collection of real-world misleading visualizations.

\section{Misviz-synth column types}
\label{sec:column_types}

Table~\ref{tab:column_types} lists the 10 column types used in the first step of the Misviz-synth synthetic visualization generation process.   Each column is assigned exactly one primary type, which determines its core data role. Additionally, columns can be assigned one or more secondary types, which describe extra attributes that help determine appropriate chart types and potential misleaders.

\begin{table}
    \centering
    \resizebox{\linewidth}{!}{ %
    \begin{tabular}{ll}
    \hline
    Column type & Characteristic\\
    \hline
    \multicolumn{2}{c}{\textit{Primary column types}} \\
    \hline
    Temporal & Contains temporal values \\
    Categorical & Contains categorical values \\
    Numerical & Contains numerical values \\
    \hline
    \multicolumn{2}{c}{\textit{Secondary column types}}  \\
    \hline
    Datetime & Datetime objects \\
    Evenly spaced, unique temporal & Evenly spaced temporal values \\
    Country & Country names \\
    Unique object & Non-numerical values appear only once \\
    Is part of whole & Numerical values sum to 1 or 100 \\
    Numerical percentage & Explicit percentage values \\
    & (e.g., marked with \%)  \\
    Potential percentage & Potential percentage values \\ 
    & (e.g., value range is [0,1] or [0,100]) \\

      \hline
    \end{tabular}}
    \caption{Primary and secondary column types used during synthetic visualization generation}
    \label{tab:column_types}
\end{table}

\section{Examples of visualizations with axis metadata}
\label{sec:axis_metadata}

Figure \ref{fig:axis_metadata} provides two examples of visualizations from Misviz-synth with the corresponding axis metadata. The latter is represented as a table and stored as a JSON file with column names as keys.

\begin{figure*}[!ht]
    \centering
\begin{tcolorbox}[colback=gray!5, colframe=gray!80, title=Task prompt for misleader detection with zero-shot MLLMs, width=\linewidth]

You are an expert in data visualization analysis. Your task is to identify misleaders present in the given visualization. \\

    Please carefully examine the visualization and detect its misleaders. Provide all relevant misleaders, up to three, as a comma separated list.
    In most cases only one misleader is relevant.
    If you detect none of the above types of misleaders in the visualization, respond with ``no misleader''. \\

    The available misleaders to select are, by alphabetical order:
    
    - discretized continuous variable: a map displays a continuous variable transformed into a categorical variable by cutting it into discrete categories, thus exaggerating the difference between boundary cases.
    
    - dual axis: there are two independent y-axis, one on the left and one on the right, with different scales.
    
    - inappropriate axis range: the axis range is too broad or too narrow.
    
    - inappropriate item order: instances of a variable along an axis are in an unconventional, non-linear or non-chronological order.
    
    - inappropriate use of line chart: a line chart is used in inappropriate or unconventional ways, e.g., using a line chart with categorical variables, or encoding the time dimension on the y-axis. 
    
    - inappropriate use of pie chart: a pie chart does not display data in a part-to-whole relationship, e.g., its shares do not sum to 100\%.
    
    - inconsistent binning size: a variable, such as years or ages, is grouped in unevenly sized bins.
    
    - inconsistent tick intervals: the ticks values in one axis are evenly spaced but their values are not, e.g., the tick value sequence is 10, 20, 40, 45.
    
    - inverted axis: an axis is displayed in a direction opposite to conventions, e.g., the y-axis displays values increasing from top to bottom or the x-axis displays values increasing from right to left.
    
    - misrepresentation:  the value labels displayed do not match the size of their visual encodings, e.g., bars may be drawn disproportionate to the corresponding numerical value.
    
    - truncated axis: an axis does not start from zero, resulting in a visual exaggeration of changes in the dependent variable with respect to the independent variable. 
    
    - 3d: the visualization includes three-dimensional effects. \\

    Provide only the final answer, without additional explanation.

\end{tcolorbox}
    \caption{Task prompt for misleader detection with zero-shot MLLMs.}
\label{fig:prompt}
\end{figure*}

\begin{figure*}[!ht]
    \centering
\begin{tcolorbox}[colback=gray!5, colframe=gray!80, title=Task prompt for misleader detection with zero-shot MLLMs, width=\linewidth]

   You are given a chart (dimensions: \{img\_dim[0]\} x \{img\_dim[1]\}) with potential misleading regions:\\
   
    Please analyze the image to detect misleaders and define bounding box coordinates for any misleading regions.
    
    ** Let’s think it step by step! ** 
    
    Here is the list of potential misleaders and their corresponding misleading regions:
    
    - discretized continuous variable: a map displays a continuous variable is transformed into a categorical variable by cutting it into discrete categories, thus exaggerating the difference between boundary cases. The misleading region is the legend of the map.
    
    - dual axis: there are two independent y-axis, one on the left and one on the right, with different scales. The misleading regions are the two vertical axes.
    
    - inappropriate axis range: the axis range is too broad or too narrow. The misleading region is the vertical axis.
    
    - inappropriate item order: instances of a variable along an axis are in an unconventional, non-linear or non-chronological order. The misleading region is (parts of) an axis.
    
    - inappropriate use of line chart: a line chart is used in inappropriate or unconventional ways, e.g., using a line chart with categorical variables, or encoding the time dimension on the y-axis.  The misleading region is one of the axis. 
    
    - inappropriate use of pie chart: a pie chart does not display data in a part-to-whole relationship, e.g., its shares do not sum to 100\%. The misleading region is the labels on the pie slices.
    
    - inconsistent binning size: a variable, such as years or ages, is grouped in unevenly sized bins. The misleading region is one of the axis or the legend for a map.
    
    - inconsistent tick intervals: the tick values in one axis are not evenly spaced, e.g., the tick value sequence is 10, 20, 40, 45. The misleading region is (parts of) one of the axis.
    
    - inverted axis: an axis is displayed in a direction opposite to conventions, e.g., the y-axis displays values increasing from top to bottom or the x-axis displays values increasing from right to left. The misleading region is one of the axis.
    
    - misrepresentation:  the value labels displayed do not match the size of their visual encodings, e.g., bars may be drawn disproportionate to the corresponding numerical value. The misleading region involves at least two objects (bar, pie slice) for which the size difference is not proportional to the value label difference.
    
    - truncated axis: an axis does not start from zero, resulting in a visual exaggeration of changes in the dependent variable with respect to the independent variable. The misleading region is the starting tick of the axis.
    - 3d: the visualization includes three-dimensional effects. The misleading region is the 3D area of the chart.\\
    
    Then output a JSON file containing coordinates for the potential misleaders and explanations.

    *** Instructions:
    
    - ** Please analyze the image (dimensions: \{img\_dim[0]\} x \{img\_dim[1]\}) to detect any misleading regions.**

    - **Provide the misleading region coordinates with the name of the corresponding misleader**
    
        - Your response format must strictly follow
        the example JSON format:
        
        ```
        [
        {{"coordinates": [[100, 200], [150, 200],[100, 300], [150, 300]],"misleader": "Truncated axis"}},
        {{"coordinates": [[250, 300], [300, 300],[250, 350], [300, 350]], "misleader": "Misrepresentation"}}]
        ```  

\end{tcolorbox}
    \caption{Task prompt variant for bounding box prediction.}
\label{fig:prompt_bbox}
\end{figure*}

\section{List of random design variations}
\label{sec:design_variations}

\textbf{All visualizations}: color variation of the background, variations of the title template, variation of font type and size, and variation of chart size.

\textbf{Bar and line charts}: addition of minor ticks in addition to major ones, positioning of the vertical axis (left or right),  variation in tick shapes, adding value labels on top of bars,  variation of the tick step size, addition of chart borders, and addition of horizontal grid lines.

\textbf{Bar charts only}: sorting bars by values or by category name, hiding the vertical axis, placing labels on top of or within the bars, using horizontal or vertical value labels, variation of bar colors.

\textbf{Line charts only}: filling area below the line with a color, variation in line style and color, and addition of horizontal or vertical grid lines.

\textbf{Pie charts only}: placing data label next to the pie slices or placing them in a legend.

\section{Zero-shot prompts}
\label{sec:prompt}

Figure \ref{fig:prompt} shows the prompt to detect misleading visualizations.  Figure \ref{fig:prompt_bbox} shows the variant used to generate bounding boxes, which is based on the prompt from \citet{chen2025unmaskingdeceptivevisualsbenchmarking}.

\begin{table*}[!ht]
  
  \centering
  \resizebox{\textwidth}{!}{ %
  \begin{tabular}{lccccccccccccc}
    \hline    
       &  \multicolumn{6}{c}{Misviz} & & \multicolumn{6}{c}{Misviz-synth} \\
         \cline{2-7}  \cline{9-14} 
    
         & Acc & Pre & Rec & F1 &  EM & PM & & Acc & Pre & Rec & F1 &  EM & PM \\
    \hline
    Image & 70.1 $\pm$ 0.9 & 73.7 $\pm$ 0.5 & 88.3 $\pm$ 1.8 & 58.7 $\pm$ 1.0 & 11.1 $\pm$ 0.7 & 14.9 $\pm$ 0.8 & & 72.0 $\pm$ 0.5 & 71.9 $\pm$ 0.9 & 92.2 $\pm$ 1.3 & 64.7 $\pm$ 1.5 & 68.0 $\pm$ 1.2 & 68.0 $\pm$ 1.2\\
    Image w. axis (DePlot\includegraphics[height=1em]{figures/fire.png}) &  72.8 $\pm$ 1.2 & 74.6 $\pm$ 0.6 & 92.0 $\pm$ 1.4 & 60.9 $\pm$ 1.4 & 12.3 $\pm$ 0.5 & 17.1 $\pm$ 0.5 & & 72.5 $\pm$ 0.7 & 72.1 $\pm$ 0.6 & 92.6 $\pm$ 0.3 & 65.2 $\pm$ 1.2 & 69.5 $\pm$ 0.9 & 69.5 $\pm$ 0.9 \\
    \hline
    
  \end{tabular}}
  \caption{Average performance with standard deviations (\%) of the image-axis classifiers on the test sets of Misviz and Misviz-synth for instances with bar, line, or pie charts.}
  \label{tab:results_classifiers_std}
\end{table*}

\section{Linter rules}
\label{sec:rules}

We explain below the linter rules for detecting some misleaders in Misviz and Misviz-synth. These rules are implemented in Python and were designed on the validation set of Misviz-synth to maximize precision over recall.

\textbf{Truncated axis} A visualization has a truncated axis if one or more of its sorted vertical axes start with a numerical value strictly higher than 0.

\textbf{Inverted axis} A visualization has an inverted axis if one or more of its sorted axes is the reverse of the default axis order.

\textbf{Dual axis} A visualization has dual axes if it has two vertical axes with different axis tick labels.

\textbf{Inappropriate item order} A visualization has an inappropriate item order if it has a temporal axis with dates shuffled in random order.  This covers only a specific case of inappropriate item order.

\textbf{Inconsistent tick intervals} A visualization has inconsistent tick intervals if the distance between the tick labels or relative positions on one or more of its numerical or temporal axes is not constant.

\textbf{Inconsistent binning size} A visualization suffers from inconsistent binning size if one axis shows categorical bins of unequal size.

\begin{table}
\resizebox{\linewidth}{!}{ %
\begin{tabular}{lcccccc}
\toprule
\multirow{2}{*}{\# Chart types} &
\multicolumn{2}{c}{Image} &
\multicolumn{2}{c}{Image w. axis (DePlot\includegraphics[height=1em] {figures/fire.png})}  & \multicolumn{2}{c}{Linter} \\
\cmidrule(lr){2-3} \cmidrule(lr){4-5} \cmidrule(lr){6-7}
 & F1 & PM & F1 & PM & F1 & PM\\
\midrule
1 (1,956 instances) & 58.2 & 15.2 & 60.7 & 17.4  & 35.3 & 7.9 \\
2 (115 instances)  & 57.2 &  6.8 & 59.5 & 11.5 & 46.3 & 6.1 \\
3 (6 instances) & 77.5 & 0.0 & 38.9 & 0.0  &  33.0 & 0.0\\
\bottomrule
\end{tabular}}
\caption{Performance of the image-axis classifiers and the linter on Misviz test (\%) as a function of the number of chart types.}
\label{tab:chart_types}
\end{table}

\begin{table}
\centering
\resizebox{\linewidth}{!}{ %
\begin{tabular}{lcccccc}
\toprule
\multirow{2}{*}{\# Chart types} &
\multicolumn{2}{c}{Image} &
\multicolumn{2}{c}{Image w. axis (DePlot\includegraphics[height=1em] {figures/fire.png})}  & \multicolumn{2}{c}{Linter} \\
\cmidrule(lr){2-3} \cmidrule(lr){4-5} \cmidrule(lr){6-7}
 & F1 & PM & F1 & PM & F1 & PM\\
\midrule
1 (1,228 instances)  & 46.9 & 13.1 & 47.9 & 14.5 & 16.8 & 7.7\\
2 (203 instances) & 46.4 & 24.3 & 48.1 & 32.7 & 17.8 & 8.9 \\
3 (10 instances) & 64.9 & 46.7 & 82.5 & 30.0 & 16.7  & 10.0 \\
\bottomrule
\end{tabular}}
\caption{Performance of the image-axis classifiers and the linter on Misviz test (\%) as a function of the number of misleader types.}
\label{tab:misleader_types}
\end{table}

\begin{table*}[!ht]
  
  \centering
  \resizebox{\textwidth}{!}{ %
  \begin{tabular}{lcccccccccccc}
    \hline    
       &  \multicolumn{3}{c}{Corpus by } & & \multicolumn{3}{c}{WTF Visualization}  & & \multicolumn{3}{c}{} 
       \\
              &  \multicolumn{3}{c}{\citet{lo2022misinformed} } & & \multicolumn{3}{c}{\citep{10670488} }  & & \multicolumn{3}{c}{\textit{r/dataisugly}} 
       \\
         \cline{2-4}  \cline{6-8}  \cline{10-12} 
    
          & Rec & EM & PM & & Rec & EM & PM & & Rec & EM & PM\\
    \hline
     Qwen2.5-VL-7B & 47.1 & 7.6 & 17.9 & & 24.7 & \digitorange{2.7} & 3.4 & & 31.8 & 7.2 & 7.2 \\
     Qwen2.5-VL-32B & 96.0 & 9.2 & 12.6 & & 96.9 & \digitorange{2.5} & 3.3 & & 93.5 & 2.7 & 2.7 \\
     Qwen2.5-VL-72B & 93.7 & 31.5 &55.5 & & 82.8 & \digitorange{20.4} & 21.5 & & 83.2 & 32.5 & 32.5 \\
     InternVL3-8B & 92.6 & \digitorange{18.1} & 37.2 & & 82.6 &   22.0&  22.7  & & 87.0 & 29.8 & 29.8 \\
     InternVL3-38B & 71.8 & 23.7 & 49.2 & & 53.9 &  34.6 & 35.5 & & 51.7 & \digitorange{21.9} & 21.9 \\
     InternVL3-78B & 80.7 & 25.4 & 46.0 & & 62.2   & 23.7 &24.3  & & 59.6 & \digitorange{21.6} & 21.6 \\
    Gemini-2.5-Flash-Lite &  86.8 & 33.1 & 60.8 & & 52.8 & \digitorange{28.0} & 28.4 & & 74.0 & \digitorange{28.0}  &29.7 \\
    GPT-4.1 & 96.2 & \digitorange{43.5} & 73.1 & & 94.2 & 58.8 & 60.2 & & 91.4 & 57.9 & 57.9 \\
    GPT-o3 & 93.1 & \digitorange{50.4} & 75.2 & & 88.9 & 63.2 & 64.2 & & 88.0 & 62.7 & 62.7 \\
    \hline
    
  \end{tabular}}
  \caption{Zero-shot MLLMs performance (\%)  on the test set of Misviz, separated by the data source of the misleading visualizations.  The data source with the lowest EM for each MLLM is marked in \digitorange{bold}.}
  \label{tab:results_per_source}
\end{table*}

\begin{figure*}[!ht]
    \centering
    \includegraphics[width=\linewidth]{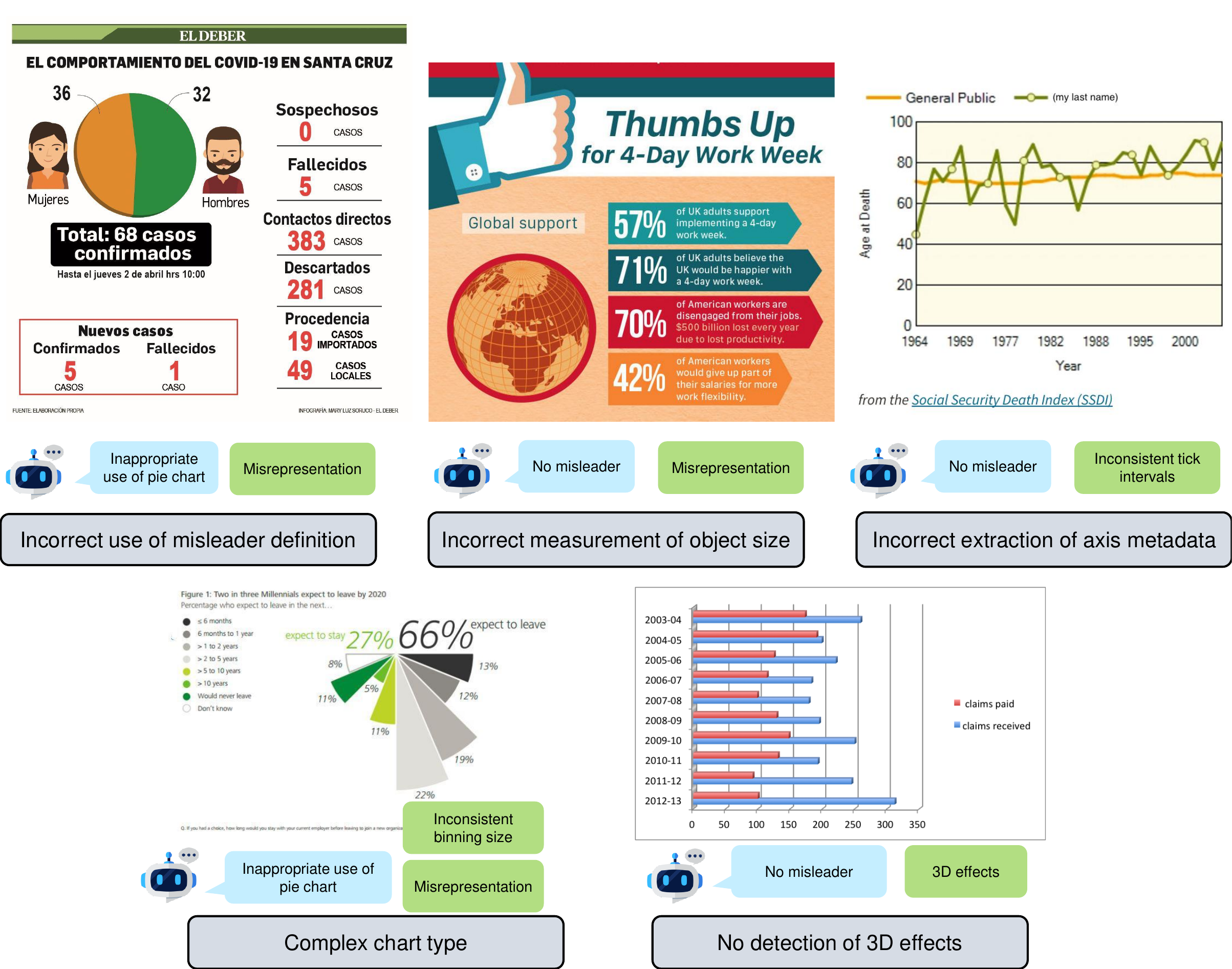}
    \caption{Error examples on the test set of  Misviz. The predictions made by GPT-4.1 are shown in blue, while the ground truth is shown in green.}
    \label{fig:error_misviz}
\end{figure*}

\section{Results with standard deviations of image-axis classifiers}
\label{sec:std_classifier}

Table \ref{tab:results_classifiers_std} reports the average results with standard deviations of the image-axis classifiers on the subset of Misviz with bar, pie, and line charts and on Misviz-synth.

\begin{figure*}[!ht]
    \centering
    \includegraphics[width=\linewidth]{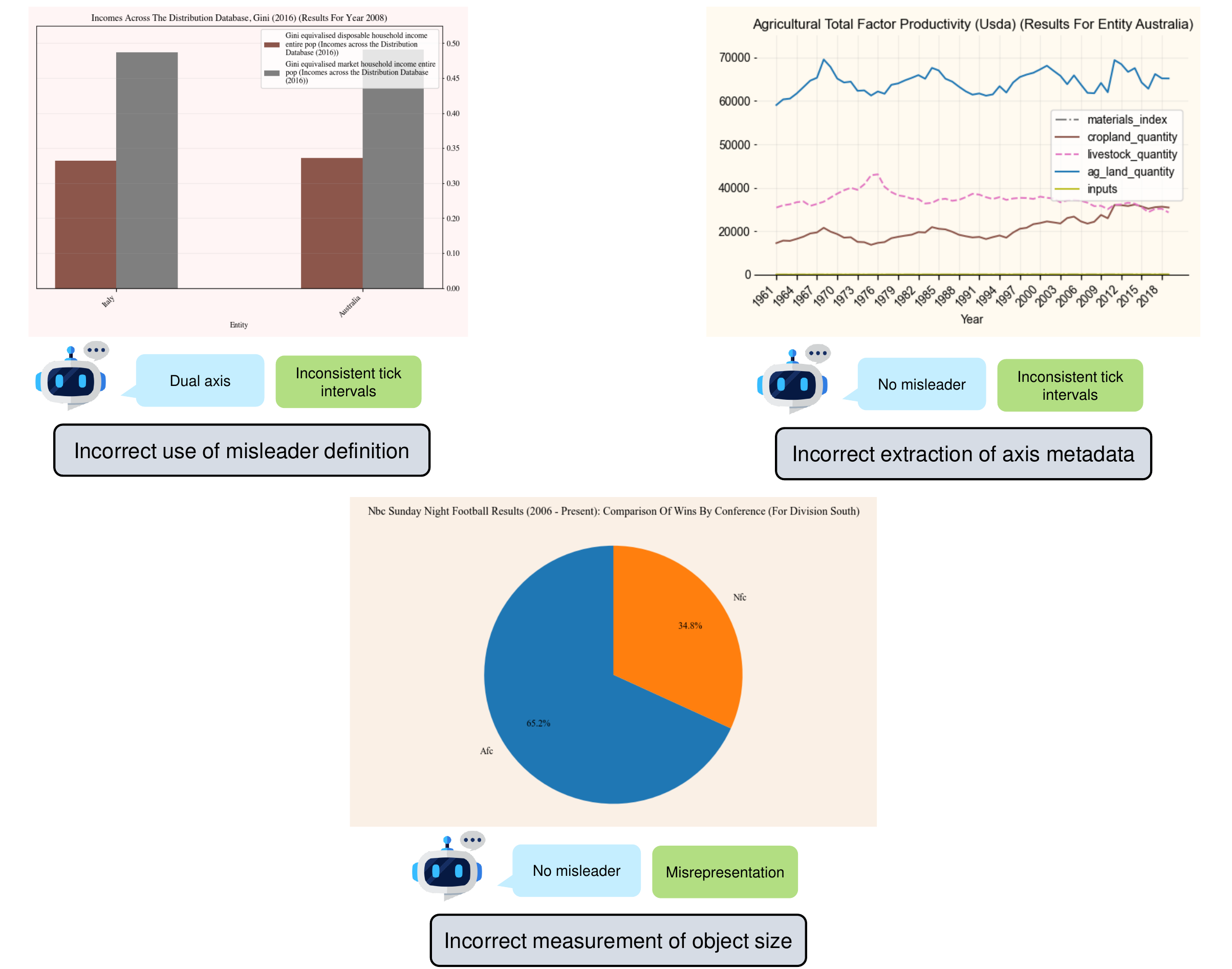}
    \caption{Error examples on the test set of  Misviz-synth. The predictions made by GPT-4.1 are shown in blue, while the ground truth is shown in green.}
    \label{fig:error_misviz_synth}
\end{figure*}

\begin{table*}[!ht]
\centering
\resizebox{\textwidth}{!}{ %
\begin{tabular}{lcccccc}
\toprule
Model &
\multicolumn{2}{c}{No definitions, no examples} &
\multicolumn{2}{c}{Definitions, no examples} &
\multicolumn{2}{c}{Definitions with one example} \\
& F1 & EM & F1 & EM & F1 & EM \\
\midrule
InternVL3-38B & 54.3 & 29.2 & 58.4 & 25.5 & 58.5 & 26.6 \\
Qwen2.5-VL-72B & 61.2 & 15.7 & 62.8 & 23.2 & 62.0 & 22.1 \\
\bottomrule
\end{tabular}}
\caption{Prompt sensitivity analysis on the val set of Misviz (\%).}
\label{tab:sensitivity_results}
\end{table*}

\section{Additional implementation details}
\label{sec:details}

\textbf{Training sets} We fine-tune DePlot on the Misviz-synth train set. The train-small set is used to train the classification head.

\textbf{DePlot fine-tuning} We fine-tune DePlot for axis metadata extraction using two H110 GPUs. We retain the original DePlot architecture and loss function \citep{liu-etal-2023-deplot}. The model is trained for 4 epochs using the Adam optimizer \citep{kingma2015adam}, a learning rate of 5e-5, and a batch size of 4. To support memory-efficient training, we apply LoRA \citep{hu2022lora} on the query and value projection layers. To enhance generalization, we apply data augmentation through random rotations and perspective transformations.

\textbf{Image-axis classifiers} The classification head is trained for up to 300 epochs with early stopping. We use a batch size of 256, a learning rate of 5e-5, and the Adam optimizer. To account for label imbalance, we apply a weighted loss based on the frequency of each misleader. The classification head has one hidden layer with 1.024 units. During training, we keep track of the best model weights. The best weights are updated at each epoch if the F1 score increases on the validation sets of both Misviz and Misviz-synth. While the classifier is not trained on Misviz, using the Misviz validation set ensures we do not overfit to the type of visualizations shown in Misviz-synth.

\section{Detailed Misviz results}
\label{sec:misviz_per_source}

Tables \ref{tab:chart_types} and \ref{tab:misleader_types} report the performance of the image-axis classifiers and the linter on the Misviz test set as a function of the number of chart types and misleader types present in the visualization, respectively. These results provide further insights into the generalization gap between the Misviz-synth training data and the real-world visualizations of Misviz. Visualizations containing multiple chart types are substantially more challenging. This is expected, as multiple chart types increase visual complexity and introduce interactions that are not represented in Misviz-synth. In contrast to chart types, increasing the number of misleaders improves performance for both classifiers and for the linter. Visualizations with more misleaders increase the likelihood that at least one misleader is detected.

Table~\ref{tab:results_per_source} reports the performance of MLLMs across three subsets of the Misviz test set, corresponding to its three sources of misleading visualizations. Performance varies significantly across subsets, but no single source is consistently easier or harder. This highlights the complementary value of each data source to the overall Misviz benchmark. Interestingly, the subset with the lowest EM is often consistent within the same model family, suggesting shared weaknesses in handling visualizations from the same data source.

\section{Prompt sensitivity analysis}
\label{sec:sensitivity_analysis}

We analyze the sensitivity of the two best-performing open-weight MLLMs to different input prompts. We consider three prompts: (1) the prompt only includes the name of the 12 misleader categories, (2) the prompt includes the definitions, (3) the default prompt, which includes the definitions and, in some cases, an example. The results for the Misviz val set, shown in Table  \ref{tab:sensitivity_results}, indicate a high level of sensitivity for EM and a moderate level for F1. The best prompt depends on the MLLM. Interestingly, InternVL3-38B achieves the highest EM when only the misleader names are provided, without any additional context.

\section{Error examples}
\label{sec:error}

Figures \ref{fig:error_misviz} and \ref{fig:error_misviz_synth} show examples of errors made by GPT-4.1  on the test sets of Misviz and Misviz-synth, respectively. One example is shown for each error category. Each example shows GPT-4.1's prediction in blue, the correct answer in green, and the error category in gray.

\end{document}